\documentclass[10pt,twocolumn]{article}

\providecommand{\tightlist}{%
  \setlength{\itemsep}{0pt}\setlength{\parskip}{0pt}}

\usepackage[margin=0.75in]{geometry}
\usepackage{dblfloatfix}
\usepackage{amsmath,amssymb}
\usepackage{graphicx}
\usepackage[hyphens,spaces,obeyspaces]{url}
\usepackage{hyperref}
\usepackage{xcolor}
\usepackage{fancyhdr}
\usepackage{titlesec}
\usepackage{booktabs}
\usepackage{longtable}
\usepackage{array}
\usepackage{calc}
\usepackage{caption}
\usepackage{enumitem}
\usepackage{fvextra}
\usepackage{framed}
\usepackage{float}

\definecolor{shadecolor}{RGB}{248,248,248}

\hypersetup{
    colorlinks=true,
    linkcolor=blue,
    citecolor=blue,
    urlcolor=blue,
    breaklinks=true,
}

\titleformat{\section}{\normalfont\large\bfseries}{\thesection}{1em}{}
\titleformat{\subsection}{\normalfont\normalsize\bfseries}{\thesubsection}{1em}{}
\titleformat{\subsubsection}{\normalfont\small\bfseries}{\thesubsubsection}{1em}{}

\setlist{nosep, leftmargin=1.5em, topsep=0.5em}
\setlist[itemize]{label=\textbullet}
\setlist[enumerate]{label=\arabic*.}

\makeatletter
\renewcommand{\maketitle}{%
\begin{center}%
{\LARGE\bfseries Meta-Benchmarks for Financial-Services LLM
Evaluation\par}%
\vspace{0.5em}%
{\large Blair Hudson\par}%
{\normalsize\itshape Commonwealth Bank of Australia\par}%
{\small\texttt{blair@cba.com.au}\par}%
\vspace{0.5em}%
{\normalsize July 2026\par}%
\end{center}%
\begin{quote}%
\small\textbf{Abstract:} Public LLM leaderboards optimise for global
average performance and do not capture the specific cognitive demands of
financial-services work: a model that leads on MMLU-Pro may underperform
on document-grounded compliance reasoning, and a coding leader may
handle multi-turn customer interactions poorly. We present a
meta-benchmarking framework that organises 452 publicly reported
benchmarks into 41 O*NET Generalized Work Activities and aggregates
those into 38 BIAN banking business domains spanning sales, operations,
risk, and support work. A multiplicative weighting scheme
(discrimination × coverage × recency), computed over a rolling model
window, rewards benchmarks that still separate the best models, are
widely reported, and remain in active use, suppressing saturated legacy
tests automatically. These weights scale the K-factor in a pairwise Elo
tournament, producing cross-benchmark-comparable work-activity scores
without raw score normalisation; business-domain scores are weighted
averages of the constituent work-activity Elos. We demonstrate the
framework on a point-in-time public snapshot covering 288 models across
25 organisations as of June 2026, and describe the methodology, full
taxonomy, design decisions, and limitations with the aim of making the
approach reproducible for institutions facing similar selection and
governance challenges.
\end{quote}%
\noindent\small\textbf{Keywords:} LLM evaluation, benchmark aggregation,
financial services, work-activity taxonomy, capability scoring,
meta-benchmarking\par%
\vspace{1em}%
}
\makeatother

\newlength{\cslhangindent}
\newlength{\csllabelwidth}
\newcommand{\citeproctext}{}
\newenvironment{CSLReferences}[2]
 {\begin{list}{}{\setlength{\leftmargin}{0pt}\setlength{\itemindent}{-\cslhangindent}\setlength{\listparindent}{\parindent}\setlength{\parsep}{3pt plus 2pt minus 1pt}}\small}
 {\end{list}}

\begin{document}
\twocolumn[{%
\maketitle
}]

\section{Introduction}\label{introduction}

Public LLM leaderboards present aggregate scores that blend performance
across tasks ranging from undergraduate trivia to competitive
programming. For an organisation selecting and governing models for
specific operational purposes, these aggregates are insufficient: a
model that leads on MMLU-Pro may underperform on document-grounded
compliance reasoning; a coding leader may score poorly on multi-turn
customer interaction tasks. Selecting and governing AI models for
regulated industries requires a more precise tool.

We describe a meta-benchmarking framework developed in a
financial-services research context to address this gap. It introduces
no new benchmarks. Instead, it aggregates 452 existing publicly reported
benchmarks into 41 work activities drawn from the O*NET Generalized Work
Activities taxonomy (National Center for O*NET Development 2024), and
further aggregates those into 38 banking business domains drawn from the
BIAN Service Landscape 14.0.0 (Banking Industry Architecture Network
2024). A dynamic weighting scheme rewards benchmarks that are actively
contested at the frontier and suppresses those that have been saturated
or are rarely attempted by recent models.

\begin{figure}[ht]
\centering
\includegraphics[width=0.62\columnwidth]{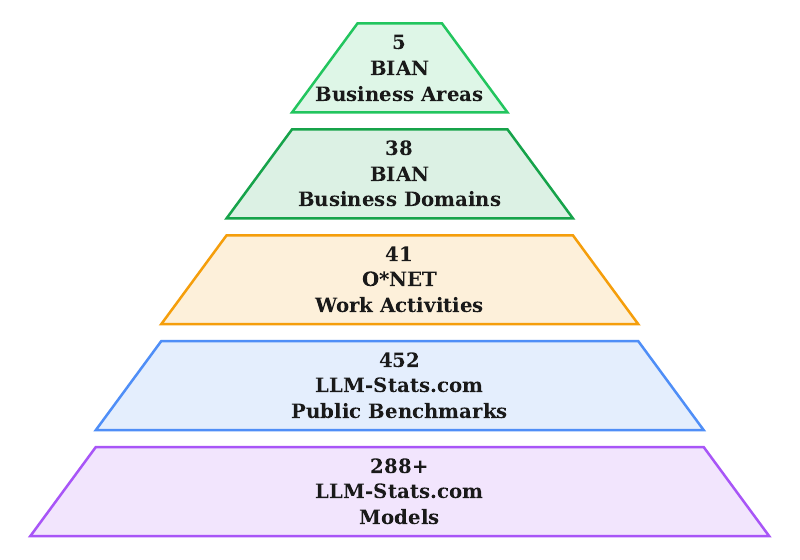}
\caption{The evaluation pyramid. Reading bottom-up, 288+ models are scored on
452 public benchmarks, which are mapped to 41 O\*NET Generalized Work
Activities, aggregated into 38 BIAN business domains, and grouped under five
BIAN Business Areas.}
\label{fig:taxonomy-funnel}
\end{figure}

The contribution is fourfold. First, we articulate a principled taxonomy
mapping benchmarks to standardised work activities and banking business
domains, grounded in established external standards (O*NET and BIAN).
Second, we describe a weighting and normalisation methodology that
handles the heterogeneous and rapidly evolving landscape of public
benchmarks without requiring manual score curation. Third, we describe
practical applications of the resulting capability profiles in
preliminary model comparison, risk-informed screening, and governance
research. Fourth, we provide sufficient methodological detail to make
the taxonomy and aggregation approach reproducible for institutions
facing similar model-selection and governance challenges.

\emph{The analysis in this paper uses publicly reported benchmark data
only. Rankings, capability assessments and all applications described
are based on this public evidence base and are not necessarily
representative of Commonwealth Bank of Australia's deployed AI systems,
operational processes, procurement decisions, or provider relationships.
Any model considered for operational use would require separate privacy,
security, legal, risk, compliance and governance review.}

\subsection{Why not evaluate on internal
benchmarks?}\label{sec:internal-challenges}

An alternative approach is to build organisation-specific benchmarks
drawn from internal data and operational workflows. This is conceptually
attractive: domain-specific evaluation should, in principle, be more
relevant than general-purpose tests. In practice, however, several
challenges make this approach hard to sustain at the current pace of
model releases.

\textbf{Velocity and coverage.} The rate of model releases has
accelerated sharply from 2024 onward (Section \ref{sec:data}). Running
internal benchmarks against each new release across every relevant
capability area can place sustained pressure on evaluation capacity in
regulated institutions.

\textbf{Execution cost.} Running even a modest benchmark suite across
hundreds of model-capability combinations requires significant API spend
or dedicated GPU allocation, a recurring cost that must be justified
against the marginal information gain relative to public benchmarks.

\textbf{Cyber and risk management overhead per model.} Internal
benchmarking requires that a model first pass supplier and cyber
onboarding, security assessment, provider due diligence, data handling
agreements, and access controls. Completing this for every candidate
before evaluation is a sequencing challenge that is hard to sustain as
the candidate pool grows, and many models are only accessible through
providers with whom the institution has no existing commercial
relationship.

\textbf{Benchmark curation and discrimination.} Constructing examples
that genuinely discriminate between frontier models, rather than being
memorised or trivially solved, requires domain expertise and iterative
refinement; a poorly designed benchmark may rank models identically and
provide no actionable signal.

\textbf{Maintenance overhead.} Financial-services task distributions
shift with new products, regulatory changes, and organisational
restructuring. Internal benchmarks require periodic re-curation to
remain valid, an ongoing cost frequently underestimated at inception.

The framework described in this paper addresses these challenges by
aggregating from the large and continuously refreshed body of public
benchmarks, applying a weighting scheme that automatically tracks which
benchmarks carry discriminating power at the current frontier, and
mapping them to a taxonomy grounded in the structure of
financial-services work. This reduces the need to use sensitive internal
data during preliminary screening, while not replacing internal
evaluation or governance.

\subsection{Methodology overview}\label{methodology-overview}

The pipeline has four stages: benchmark collection and normalisation
(Section \ref{sec:data}), composite weight computation (Section
\ref{sec:weighting}), pairwise Elo scoring (Section \ref{sec:tasks}),
and capability aggregation (Section \ref{sec:capabilities}).

\begin{figure}[ht]
\centering
\includegraphics[width=\columnwidth]{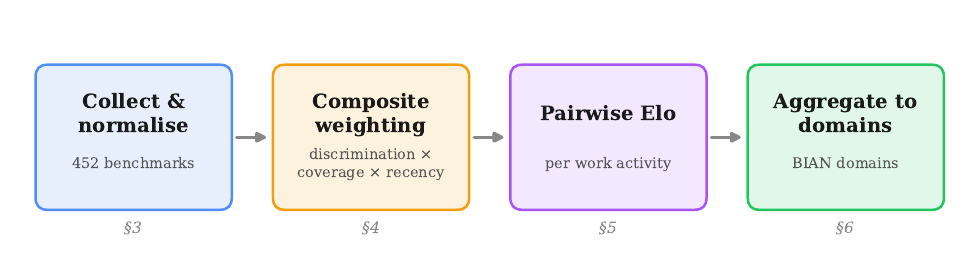}
\caption{The four-stage pipeline: benchmarks are collected and normalised,
assigned composite weights from discrimination, coverage and recency, scored
into a per-work-activity Elo rating through pairwise comparison, and aggregated
into BIAN business-domain profiles.}
\label{fig:pipeline}
\end{figure}

Pairwise Elo treats each benchmark as a series of head-to-head model
comparisons: for every pair of models sharing a benchmark score, the
higher-scoring model is credited with a win and the lower with a loss,
and each comparison updates their ratings by an amount proportional to
the benchmark's composite weight. Running these comparisons across all
benchmarks in a task produces a single rating scale comparable across
benchmarks regardless of their absolute score ranges or difficulty. The
mechanics are detailed in Section \ref{sec:tasks}.

\section{Related Work}\label{related-work}

\subsection{General Evaluation and Benchmark
Aggregation}\label{general-evaluation-and-benchmark-aggregation}

Aggregating LLM performance into composite scores has been approached
from several angles. The Holistic Evaluation of Language Models (HELM)
framework (Liang et al. 2023) evaluates models on a curated suite of
scenarios and metrics, emphasising coverage and transparent metric
construction; BIG-Bench and its harder variant (Srivastava et al. 2023;
Suzgun et al. 2023) aggregate over 200 contributed tasks; and the Open
LLM Leaderboard (Fourrier et al. 2024) maintains continuous community
rankings with periodic refreshes to counteract saturation. The
lm-evaluation-harness (Gao et al. 2024) standardises execution for
reproducing published results. Our work is complementary: we aggregate
scores from public reports rather than rerunning evaluations, focusing
on the meta-level question of which benchmarks carry evidential weight
for a given domain.

Chatbot Arena (Chiang et al. 2024) applies pairwise Elo over human
preference judgements, providing a conversational-quality measure
orthogonal to task-based benchmarks. We use a similar pairwise
comparison procedure, but over reported benchmark scores, normalised to
a 50-centred scale for interpretability (Section \ref{sec:tasks}).

\subsection{Benchmark Saturation, Contamination, and Dynamic
Evaluation}\label{benchmark-saturation-contamination-and-dynamic-evaluation}

A key challenge in benchmark aggregation is saturation: once a critical
mass of models exceeds 90\% accuracy on a benchmark, it ceases to
discriminate among frontier systems (Kiela et al. 2021). MMLU (Hendrycks
et al. 2021), for example, was the dominant general benchmark through
2023 but saturated as frontier models consistently exceeded 90\%
accuracy. MMLU-Pro (Y. Wang et al. 2024) was designed as a harder
successor with more reasoning- intensive questions and greater
resistance to prompting artefacts. GPQA (Rein et al. 2024) targets
expert-level science questions, and the Humanity's Last Exam (HLE) (Phan
et al. 2025) comprises questions that evaded solution by all tested
models at release. LiveBench (White et al. 2025) was designed explicitly
around contamination concerns: it is updated monthly with new problems
drawn from recent competitive programming, mathematics, and reasoning
challenges, making it difficult to include in pre-training data.

Contamination risks, where benchmark examples appear in pre-training
corpora, compound the saturation problem (Xu et al. 2024; Chen et al.
2025). Our weighting scheme addresses both concerns by measuring
discrimination among top scorers within a rolling time window,
automatically suppressing saturated or contaminated benchmarks as their
frontier-separation falls without requiring manual benchmark lifecycle
management.

\subsection{Task Taxonomies and Capability
Ontologies}\label{task-taxonomies-and-capability-ontologies}

Grouping NLP tasks into higher-level categories has a long history.
Earlier task-suite benchmarks such as GLUE (A. Wang, Singh, et al. 2019)
and SuperGLUE (A. Wang, Pruksachatkun, et al. 2019) established the
practice of grouped evaluation for natural language understanding,
subsequently superseded by larger aggregate benchmarks as model
capability grew. In contrast to these supply-side groupings, we define
the evaluation taxonomy from the demand side, grounded in two
established external standards: the O*NET Generalized Work Activities
catalogue of the cognitive work commonly performed across occupations,
and the BIAN Service Landscape of banking business domains, rather than
the set of available benchmarks.

\subsection{Finance and Enterprise LLM
Evaluation}\label{finance-and-enterprise-llm-evaluation}

General leaderboards do not reliably predict performance on
domain-specific tasks. Wu et al. (2023) show that a finance-adapted
model outperforms larger general models on financial NLP tasks,
establishing that domain specificity matters for evaluation as well as
training. FinBen (Q. Xie et al. 2024) evaluates 42 financial tasks
across 24 dimensions, and FinanceBench (Islam et al. 2023) reports that
even state-of-the-art models with retrieval incorrectly answer or refuse
a large share of questions over real company filings, a finding that
directly motivates domain-specific evaluation. HELM Finance (Stanford
CRFM 2024) adapts HELM to financial tasks including FinQA, FinanceBench,
and BANKING77. Collectively these works establish that domain-specific
evaluation matters and that general leaderboards are insufficient for
regulated-industry deployment decisions.

\subsection{Agentic and Workflow
Evaluation}\label{agentic-and-workflow-evaluation}

Financial-services use cases increasingly involve agentic workflows:
multi-turn interactions with tools, structured data sources, and
external systems. \(\tau\)-bench (Yao et al. 2025) evaluates agents
interacting with tools and simulated users in realistic domains while
following domain-specific rules, reporting that even strong models
struggle with reliable task completion in customer-service settings.
SWE-bench Verified (OpenAI 2024) is a human-validated subset of software
engineering tasks providing reliable coding-agent evaluation directly
relevant to software-engineering work activities. The Berkeley Function
Calling Leaderboard (Patil et al. 2023) evaluates function and
tool-calling reliability on real-world APIs and is periodically updated,
making it a key discriminator in our Tool Use task. OSWorld (T. Xie et
al. 2024) provides execution-based evaluation of multimodal computer-use
agents on real desktop environments, directly relevant to autonomous
operational tasks in the Vision \& GUI dimension. These agentic
benchmarks complement knowledge-based tests by evaluating reliable
execution rather than accuracy on isolated questions.

\section{Benchmark Collection}\label{sec:data}

\subsection{Data Source}\label{data-source}

Benchmark scores are sourced from the LLM Stats API ({``{LLM Stats}:
{A}ggregated {LLM} Benchmark Results''} 2024)
(\url{https://llm-stats.com}), which aggregates publicly reported
evaluation results from provider technical reports, third-party
evaluation organisations, and community-maintained leaderboards.

Scores are aggregated into a single versioned snapshot: one record per
model, holding its raw benchmark scores, benchmark metadata (label, date
range, score distribution), and the benchmark-to-activity and
activity-to-domain mappings described in subsequent sections. Raw
upstream scores are not modified; work-activity and business-domain
scores are derived from them using the methodology that follows.

\textbf{Data snapshot.} All scores in this paper are drawn from a fetch
performed in June 2026. The snapshot is pinned at this date; scores for
subsequent model releases or retroactive benchmark additions are not
reflected, and results may differ if regenerated against a later
snapshot.

\subsection{Model Coverage}\label{model-coverage}

As of June 2026, the dataset covers 288 models from 25 organisations
spanning the period January 2022 to June 2026. Organisations represented
include Anthropic, Google DeepMind, OpenAI, Meta AI, Mistral, xAI,
Alibaba (Qwen), Cohere, and others. Models range from sub-7B parameter
open-weight instruction models to proprietary frontier systems. The
dataset includes both open-weight models (released under Apache 2.0,
MIT, Llama Community, and similar licences) and proprietary API-access
models.

\subsection{Model Releases}\label{model-releases}

\begin{figure}[ht]
\centering
\includegraphics[width=\columnwidth]{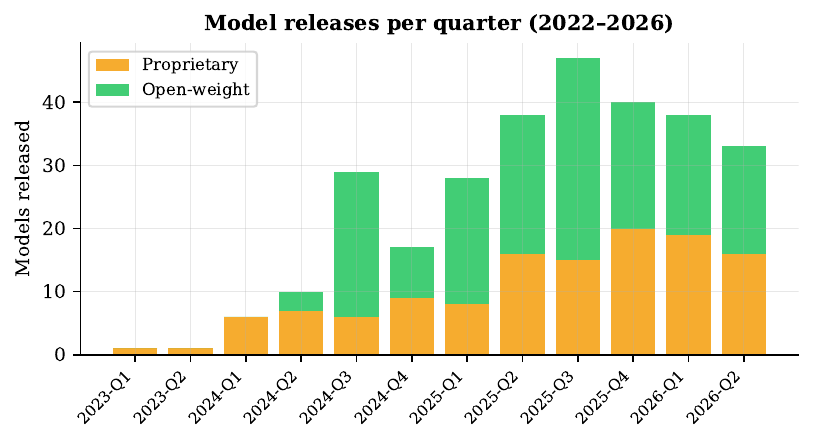}
\caption{Model releases per quarter (2022--2026), split by open-weight (green)
and proprietary (amber). Release pace accelerated markedly from 2024-Q2
onward.}
\label{fig:model-releases}
\end{figure}

The count of models released per quarter, broken down by open-weight and
proprietary type, is shown in Figure \ref{fig:model-releases}. Release
pace accelerated markedly from 2024 onward within the indexed model
population.

\subsection{Benchmark Catalogue}\label{benchmark-catalogue}

The current catalogue contains 452 unique benchmark identifiers mapped
across 41 work activities (Section \ref{sec:taxonomy}). The mapping is
many-to-many: a benchmark may appear in multiple work activities (for
example, MMLU maps to Analyzing Data or Information, Judging the
Qualities of Objects, Services, or People, and Updating and Using
Relevant Knowledge), and each work activity contains multiple
benchmarks. The total number of benchmark--work-activity assignments is
1\{,\}750.

Benchmarks with fewer than three scored models in the current rolling
window receive weight zero and do not contribute to work-activity scores
for that period; they remain in the catalogue for historical periods.
The pipeline does not impute missing scores, a model without a score on
a benchmark is not included in scoring for that benchmark.

The number of benchmark identifiers per work activity is shown in Figure
\ref{fig:benchmark-counts}; the full list is provided in Appendix
\ref{sec:appendix}.

\begin{figure}[ht]
\centering
\includegraphics[width=\columnwidth]{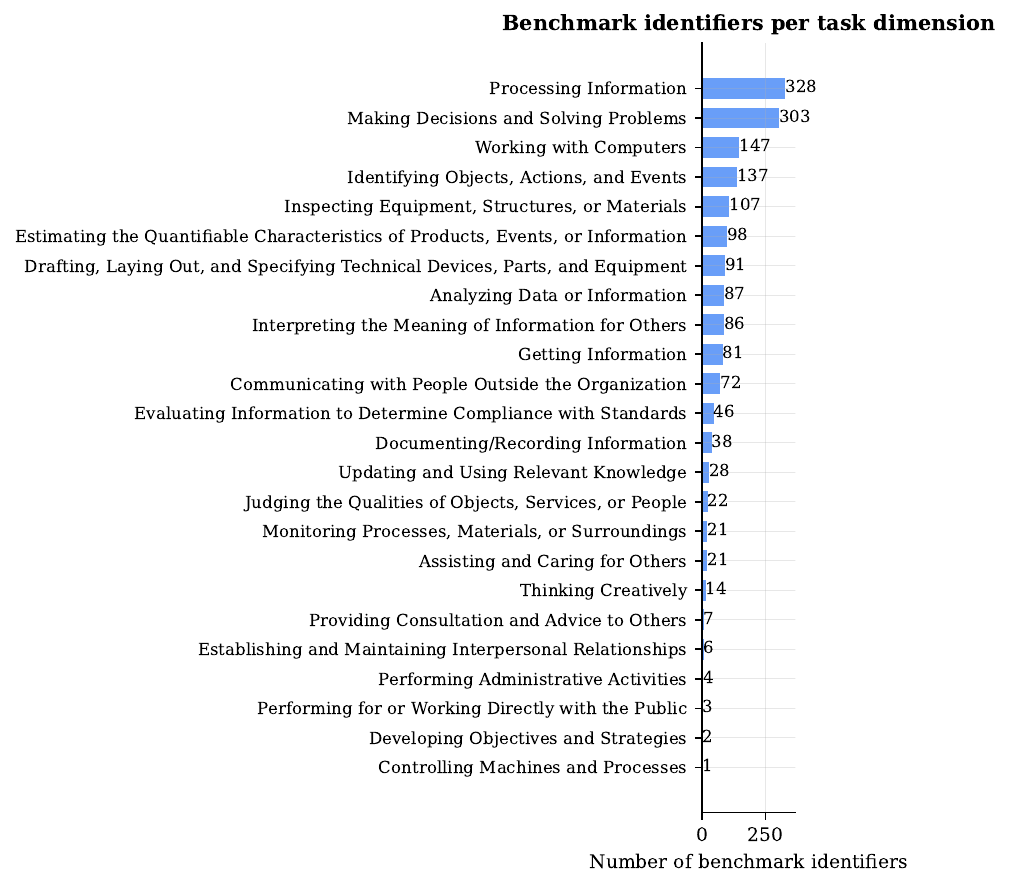}
\caption{Number of benchmark identifiers assigned to each work activity.
Information-processing activities such as Processing Information and Making
Decisions and Solving Problems are the most heavily covered; physical and
managerial activities have no public benchmark coverage.}
\label{fig:benchmark-counts}
\end{figure}

\section{Task Groupings and Composite Benchmark
Weights}\label{sec:weighting}

\subsection{Benchmark-to-Task Mapping}\label{benchmark-to-task-mapping}

Each benchmark is assigned to one or more task dimensions via a
structured configuration file. The mapping is managed independently of
the pipeline code so new benchmarks can be categorised without code
changes. Task assignments are made on the basis of what the benchmark
primarily measures, not the domain of its questions: for example, MMLU
appears in both General Reasoning and Finance \& Business because its
questions span both domains; SWE-Bench appears only in Coding because it
tests software engineering task completion.

When a benchmark appears in multiple tasks, its score contributes to the
computation of each assigned task independently. There is no
normalisation across tasks to prevent double-counting, a benchmark like
MMLU appearing in three tasks will influence all three task scores
separately.

\subsection{Composite Benchmark
Weights}\label{composite-benchmark-weights}

Not all benchmarks contribute equally to a task score. A benchmark on
which all evaluated models score above 90\% carries no information about
relative capability. Conversely, a benchmark that only 5\% of recent
models have attempted may reflect specialisation bias rather than broad
model quality. The framework addresses both concerns with a
multiplicative weight:

\[w_b = \text{disc}_b \times \text{coverage}_b \times \text{recency}_b\]

In plain terms, a benchmark earns weight only if it still separates the
top models (discrimination), is widely reported (coverage), and is still
being attempted by recently released models (recency); failing any one
of the three heavily discounts it. Each factor is independently
normalised to \([0, 1]\) across all active benchmarks in the current
period, so that the best benchmark on each dimension scores 1.0.
Normalising across all benchmarks rather than within each task means a
benchmark's weight reflects its absolute discriminative power relative
to the full population: a benchmark in a task with strong frontier
competition receives a higher weight than one in a task with uniformly
weak separation, even if both rank first within their own task group.
The three factors are:

\textbf{Discrimination} (\(\text{disc}_b\)): a benchmark is only useful
for ranking if the best models still score differently on it. We measure
this as the spread of scores among the top 10 scorers on benchmark \(b\)
from models released within the rolling window (default: 12 months): a
wide spread means the benchmark still separates the leaders, while a
narrow spread means they have all approached the ceiling. Restricting to
the top 10 ensures that discrimination reflects separation at the
frontier rather than variance introduced by weaker models that have no
bearing on which leading model is best. Benchmarks with fewer than 3
recent models receive \(\text{disc}_b = 0\).

\textbf{Coverage} (\(\text{coverage}_b\)): the fraction of all models
released within the rolling window that have a score on benchmark \(b\).
A benchmark that only 5\% of recent models have attempted carries weaker
evidence than one that is near-universal. Coverage is best interpreted
as a confidence weight rather than a validity weight: a widely-reported
benchmark is not necessarily better, but there is stronger evidence that
the score is representative. A niche benchmark highly relevant to
financial services (e.g.~a compliance-specific test with few reported
results) will receive lower coverage weight; this is a known limitation
addressed in Section \ref{sec:limitations}.

\textbf{Recency} (\(\text{recency}_b\)): the mean normalised age of
recent model scores on benchmark \(b\), where 0 represents the oldest
model in the window and 1 represents the newest. Benchmarks where the
most recently released models have scores are preferred over those that
only older models in the window have attempted.

\subsection{Weight Formula, Properties, and Benchmark
Lifecycle}\label{sec:lifecycle}

The multiplicative form ensures a benchmark must perform well on all
three dimensions to receive significant weight. A highly discriminating
benchmark that is rarely attempted (low coverage) or only by old models
(low recency) is down-weighted accordingly. For example, a benchmark
with \(\text{disc} = 0.9\), \(\text{coverage} = 0.2\),
\(\text{recency} = 0.3\) receives weight
\(0.9 \times 0.2 \times 0.3 = 0.054\), nearly irrelevant despite high
discrimination.

The weight \(w_b\) scales the influence of benchmark \(b\) on task
scores; its role in the scoring algorithm is described in Section
\ref{sec:tasks}.

Weights are computed \textbf{per period}: at each point in the timeline,
only models available up to that month are used to compute
discrimination, coverage, and recency. Early periods (2022--2023) are
therefore governed by the benchmarks that were discriminating then
(e.g.~MMLU, HellaSwag), while recent periods up-weight newer harder
tests (e.g.~GPQA, HLE, ARC-AGI-v2). This also handles the benchmark
lifecycle without manual intervention: as frontier models converge on a
benchmark and its discrimination falls toward zero, its composite weight
approaches zero regardless of coverage or recency. The benchmark stops
contributing to scores even though its historical results remain. This
pattern, active discrimination, saturation, supersession, is well
documented in the evaluation literature (Liang et al. 2023; Kiela et al.
2021).

\begin{figure}[ht]
\centering
\includegraphics[width=\columnwidth]{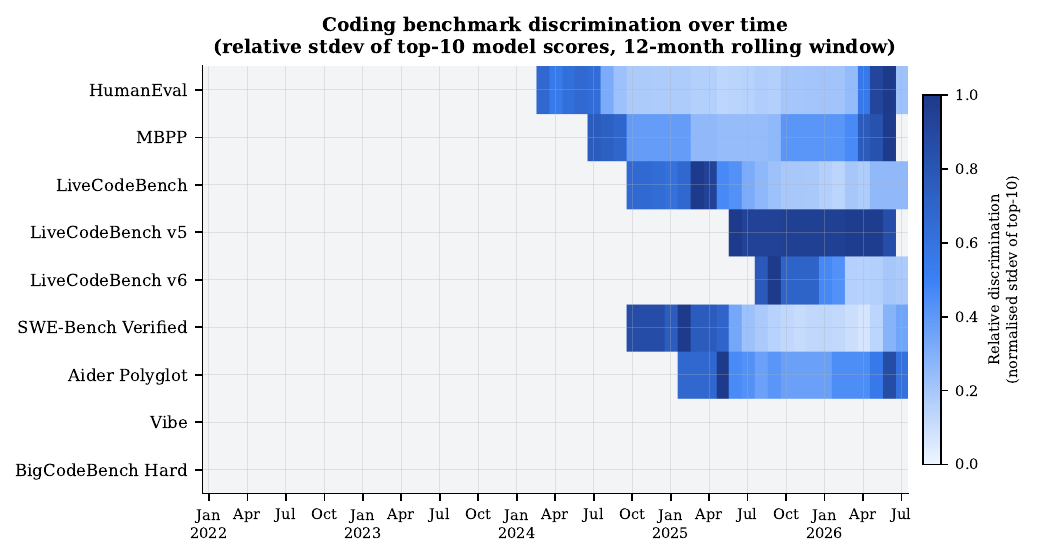}
\caption{Discrimination heat map for selected coding benchmarks, 2022--2026.
Cell intensity represents the normalised standard deviation of top-10 model
scores within the 12-month rolling window ending at each month (grey = no
data; dark blue = maximum discrimination). HumanEval and MBPP saturate early;
LiveCodeBench variants and agentic benchmarks (SWE-Bench, Aider) sustain
frontier separation through 2026.}
\label{fig:coding-timeline}
\end{figure}

This lifecycle pattern is illustrated for a curated set of coding
benchmarks from 2022 to 2026 in Figure \ref{fig:coding-timeline}.
HumanEval and MBPP, introduced as foundational code generation tests,
show strong initial discrimination that erodes progressively as models
achieve near-perfect pass rates. LiveCodeBench variants maintain higher
discrimination as the benchmark is designed for continuous refresh with
new competitive programming problems. SWE-Bench Verified and Aider
Polyglot introduce agentic coding evaluation from 2024 onward, opening a
new discrimination frontier as models begin to differ substantially in
their ability to complete real-world software tasks rather than isolated
function generation.

\section{Aggregate Task Score}\label{sec:tasks}

\subsection{Pairwise Elo}\label{sec:normalization}

Raw benchmark scores are not directly comparable across benchmarks. GPQA
scores cluster in the range 0.5--0.8 for frontier models while HumanEval
scores span 0.4--0.99; AIME 2025 scores for leading models span
0.6--0.95 while HellaSwag scores approach a ceiling near 0.95 for the
same models. A weighted average of raw scores would mix benchmark
difficulty, scale, and spread into a single number that means little.

We address this using \textbf{pairwise Elo}: for each task \(T\), models
are compared head-to-head on every benchmark assigned to that task. The
composite benchmark weight \(w_b\) from Section \ref{sec:weighting}
scales the K-factor for that benchmark's pairwise comparisons.
Benchmarks with high discrimination, coverage, and recency exert
proportionally more influence over the final rating; saturated or
low-coverage benchmarks exert proportionally less.

For a pair of models \((i, j)\) compared on benchmark \(b\):

\[K_b = K_{\text{base}} \cdot w_b \qquad \Delta R_i = K_b \bigl(S_i - E_i\bigr)\]

where \(K_{\text{base}} = 32\), \(S_i \in \{0, 0.5, 1\}\) is the match
outcome (win/draw/loss determined by raw benchmark score), and
\(E_i = 1/(1 + 10^{(R_j - R_i)/400})\) is the expected score under the
current ratings, that is, the win probability implied by the rating gap.
A model gains rating when it wins more than expected and loses rating
when it wins less. The K-factor controls how much a single comparison
can move a rating, so scaling it by the benchmark weight \(w_b\) lets
stronger benchmarks pull ratings harder than saturated or
rarely-attempted ones. Ratings are initialised at 1500 (equal for all
models). The tournament runs for 20 passes over all model-pairs on all
benchmarks in the task. Multiple passes are needed because early
comparisons in a pass shift the ratings used by later comparisons, so
information must propagate over several iterations; by 20 passes the
per-pass changes are negligible and ratings are treated as converged.

Models absent from a benchmark contribute no comparisons for that
benchmark: they are neither rewarded for absence nor penalised. Their
Elo is determined entirely by the benchmarks they did attempt. Section
\ref{sec:validation} examines evidence density per capability
explicitly.

\subsection{Normalisation and Scale}\label{normalisation-and-scale}

After convergence, per-task ratings are normalised across all
participating models to a fixed mean of 50 and standard deviation of 10:

\[\text{Elo}_m^{(T)} = 50 + 10 \cdot \frac{R_m - \bar{R}}{\sigma_R}\]

where \(R_m\) is model \(m\)'s converged pairwise rating for task \(T\),
and \(\bar{R}\), \(\sigma_R\) are the mean and standard deviation across
all participating models. In effect this restates each model's rating as
a distance from the population average in standard-deviation units, then
rescales so that 50 is the mean and each 10 points is one standard
deviation. This makes Elo scores comparable across tasks: 50 always
represents the population mean, and a model two standard deviations
above average scores Elo 70 regardless of the number of benchmarks in
the task or their absolute score ranges.

To read a score concretely: Elo 50 is exactly the average model on that
task; Elo 60 is one standard deviation above average (roughly the top
sixth of the field); Elo 70 is two standard deviations above (roughly
the top few per cent); and Elo 40 is one standard deviation below
average. A 10-point gap between two models therefore always means the
same thing, a full standard deviation of separation, whether the task
has three benchmarks or thirty.

Scores are precomputed by the data pipeline and stored in the model
records. Because Elo is recomputed from scratch on each run, scores may
shift as new benchmarks and models are added, so the score always
reflects current comparative evidence rather than a frozen historical
snapshot (the cross-run stability implications are discussed in Section
\ref{sec:limitations}).

For timeline charts, each period \(t\) selects the best proprietary
model and best open-weight model from all models released up to that
month. There is no monotonic carry-forward of scores, if a newer model
surpasses a previous leader, the old leader is simply no longer the
period's representative.

\begin{figure}[ht]
\centering
\includegraphics[width=\columnwidth]{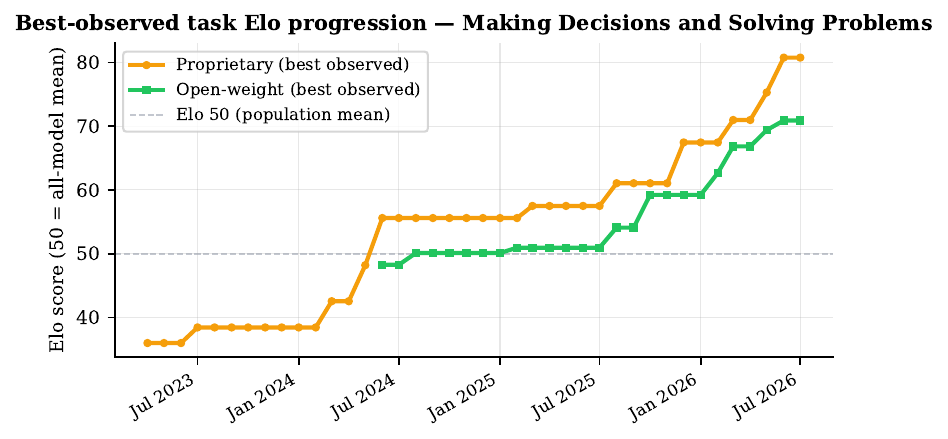}
\caption{Best-observed Elo score progression for the Making Decisions and Solving
Problems work activity in this dataset,
January 2023 -- June 2026. Amber: best proprietary model in each period. Green:
best open-weight model. The gap narrows as open-weight models approach frontier
performance.}
\label{fig:timeline-general}
\end{figure}

\begin{figure}[ht]
\centering
\includegraphics[width=\columnwidth]{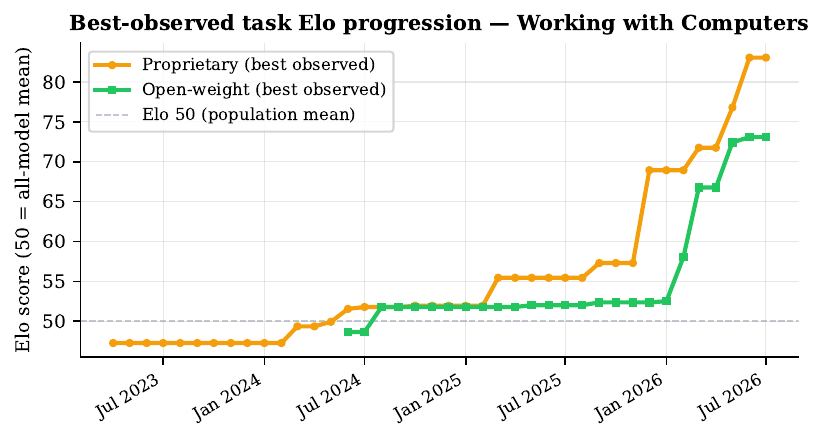}
\caption{Best-observed Elo score progression for the Working with Computers work activity.
The open-weight trajectory shows step-change improvements corresponding to major
model releases in this benchmark snapshot.}
\label{fig:timeline-coding}
\end{figure}

Figures \ref{fig:timeline-general} and \ref{fig:timeline-coding}
illustrate the best-observed Elo score trajectory for the Making
Decisions and Solving Problems and Working with Computers work
activities respectively, from January 2023 to June 2026. The gap between
proprietary (amber) and open-weight (green) best-observed scores narrows
substantially over this period across both task dimensions.

\begin{figure}[ht]
\centering
\includegraphics[width=\columnwidth]{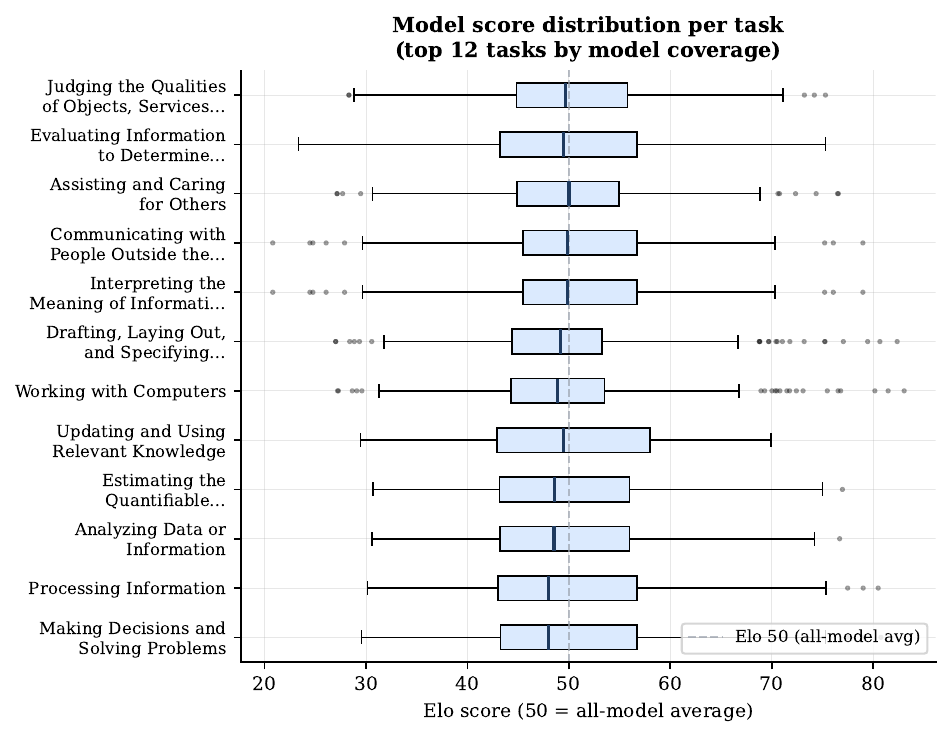}
\caption{Distribution of model Elo scores per task (top 12 tasks by model
coverage). The vertical dashed line marks Elo 50 (all-model average). Wide
interquartile ranges confirm that benchmarks continue to discriminate across
the full 288-model population.}
\label{fig:score-distribution}
\end{figure}

The distribution of Elo scores across all 288 models for the 12
most-covered work activities is shown in Figure
\ref{fig:score-distribution}. Wide interquartile ranges confirm that
current benchmarks continue to meaningfully differentiate models within
the dataset.

\section{Capability Scores}\label{sec:capabilities}

The capability taxonomy is a research abstraction for organising public
benchmark evidence; it is not a statement of deployed systems, approved
use cases, or production readiness (see Section
\ref{sec:score-semantics}).

\subsection{Two-Layer Taxonomy}\label{two-layer-taxonomy}

The framework grounds both of its abstraction layers in established
external standards. The task layer uses the \textbf{O*NET Generalized
Work Activities} (GWAs): 41 broad categories of human work activity
maintained by the U.S. Department of Labor, each defined by a set of
Intermediate Work Activities that describe the activity in detail. The
capability layer uses the \textbf{BIAN Service Landscape 14.0.0}
business domains: 38 standardised banking business domains grouped into
five Business Areas. Both standards are versioned, externally
maintained, and independent of any single institution, which makes the
taxonomy reproducible and resistant to the selection bias of custom
taxonomies.

Public benchmarks are mapped to GWAs (the task layer), and GWAs are
mapped to BIAN business domains (the capability layer). The mapping is
many-to-many at both layers: a single benchmark may exercise several
work activities, and a single work activity contributes evidence to
several business domains.

\subsection{BIAN Business Domains}\label{bian-business-domains}

The 38 business domains are organised into the five BIAN Business Areas:

\begin{itemize}
\item \textbf{Sales and Service}: Channel Specific, Cross Channel, Marketing,
Sales, Customer Management, Servicing.
\item \textbf{Reference Data}: Party, External Agency, Market Data, Product
Management.
\item \textbf{Operations and Execution}: Product Specific Fulfillment, Loans and
Deposits, Investment Management, Trade Banking, Wholesale Trading, Cards, Market
Operations, Corporate Financing and Advisory Services, Consumer Services, Cross
Product Operations, Payments, Account Management, Operational Services,
Collateral Administration.
\item \textbf{Risk and Compliance}: Bank Portfolio and Treasury, Business
Analysis, Regulations and Compliance, Models.
\item \textbf{Business Support}: IT Management, Non-IT and Non-HR Enterprise
Services, Buildings Equipment and Facilities, Business Command and Control,
Finance, Human Resource Management, Knowledge and Intellectual Property
Management, Corporate Relations, Business Direction, Document Management and
Archive.
\end{itemize}

\subsection{Taxonomy Derivation}\label{taxonomy-derivation}

Each public benchmark is first mapped to one or more O*NET GWAs based on
the cognitive work it actually exercises, read against the GWA
definitions and their Intermediate Work Activities. A coding benchmark,
for example, maps to \emph{Working with Computers} and
\emph{Drafting, Laying Out, and Specifying
Technical Devices}; a factual-recall benchmark maps to \emph{Getting
Information} and
\emph{Evaluating Information to Determine Compliance with
Standards}. Each GWA is then mapped to the BIAN business domains whose
work it informs, so that benchmark evidence flows upward from benchmarks
to work activities to business domains.

All 41 GWAs are mapped to at least one business domain, so every domain
is defined in the taxonomy. Benchmark coverage, however, is uneven: only
24 of the 41 GWAs are exercised by any public benchmark. The remaining
17 are physical and managerial activities (e.g.~Performing General
Physical Activities, Operating Vehicles, Coaching and Developing Others,
Resolving Conflicts and Negotiating, Staffing) for which no public
language-model benchmark provides evidence. We retain them in the
taxonomy rather than dropping them, because their absence is itself a
finding: the public benchmark ecosystem concentrates on
information-processing work and is largely silent on the physical and
interpersonal activities that make up much of real-world banking work.

The configuration files that define both mapping layers are versioned
alongside the data snapshot (Section \ref{sec:data}), and are updated as
new benchmarks emerge for underserved activities. The mappings presented
here are a candidate set rather than a definitive one; their status and
the rationale for aligning to external standards rather than inventing a
bespoke taxonomy are discussed in Section \ref{sec:conclusion}.

\subsection{Score Semantics}\label{sec:score-semantics}

A business-domain Elo score reflects the weight of publicly available
benchmark evidence, filtered through the GWA and business-domain
mapping, but does not substitute for:

\begin{itemize}
  \item internal evaluation against production task distributions;
  \item privacy, data governance, and security review;
  \item model risk assessment and explainability review;
  \item human-factors and accessibility evaluation;
  \item legal review of model licence and terms of use; or
  \item regulatory compliance assessment against applicable obligations.
\end{itemize}

The framework is designed to support the
\emph{model selection and screening} phase of the model assessment
process, providing a structured, reproducible basis for comparing a
large population of models before committing to deeper evaluation, not
to replace the validation that follows selection.

Business domains whose only contributing GWAs are uncovered by public
benchmarks inherit no benchmark evidence and therefore carry no score.
They remain in the taxonomy so that coverage gaps are visible rather
than hidden.

\subsection{Capability Aggregation
Formula}\label{capability-aggregation-formula}

A business-domain score is the weighted average of the Elo scores of its
constituent work activities:

\[\text{Elo}_\text{cap} = \frac{\displaystyle\sum_{t \in C} w_t \cdot \text{Elo}_{m,t}}{\displaystyle\sum_{t \in C} w_t}\]

where \(C\) is the set of work activities contributing to the business
domain and \(w_t\) is the weight assigned to work activity \(t\) within
that domain. By default, all work activities within a domain contribute
equally (\(w_t = 1\) for all \(t \in C\)), making the default formula a
simple unweighted average of work-activity Elo scores. Weights are
normalised internally (they do not need to sum to a specific value) so
that the meaningful quantity is the relative weight of one work activity
against another.

Because the constituent work-activity scores use dynamically computed
benchmark weights (Section \ref{sec:weighting}), business-domain scores
are self-updating: as the model population evolves, newly discriminating
benchmarks gain weight and saturated ones lose it automatically (Section
\ref{sec:lifecycle}).

\subsection{Work-Activity--to--Domain Mapping}\label{sec:taxonomy}

\begin{figure}[ht]
\centering
\includegraphics[width=\columnwidth]{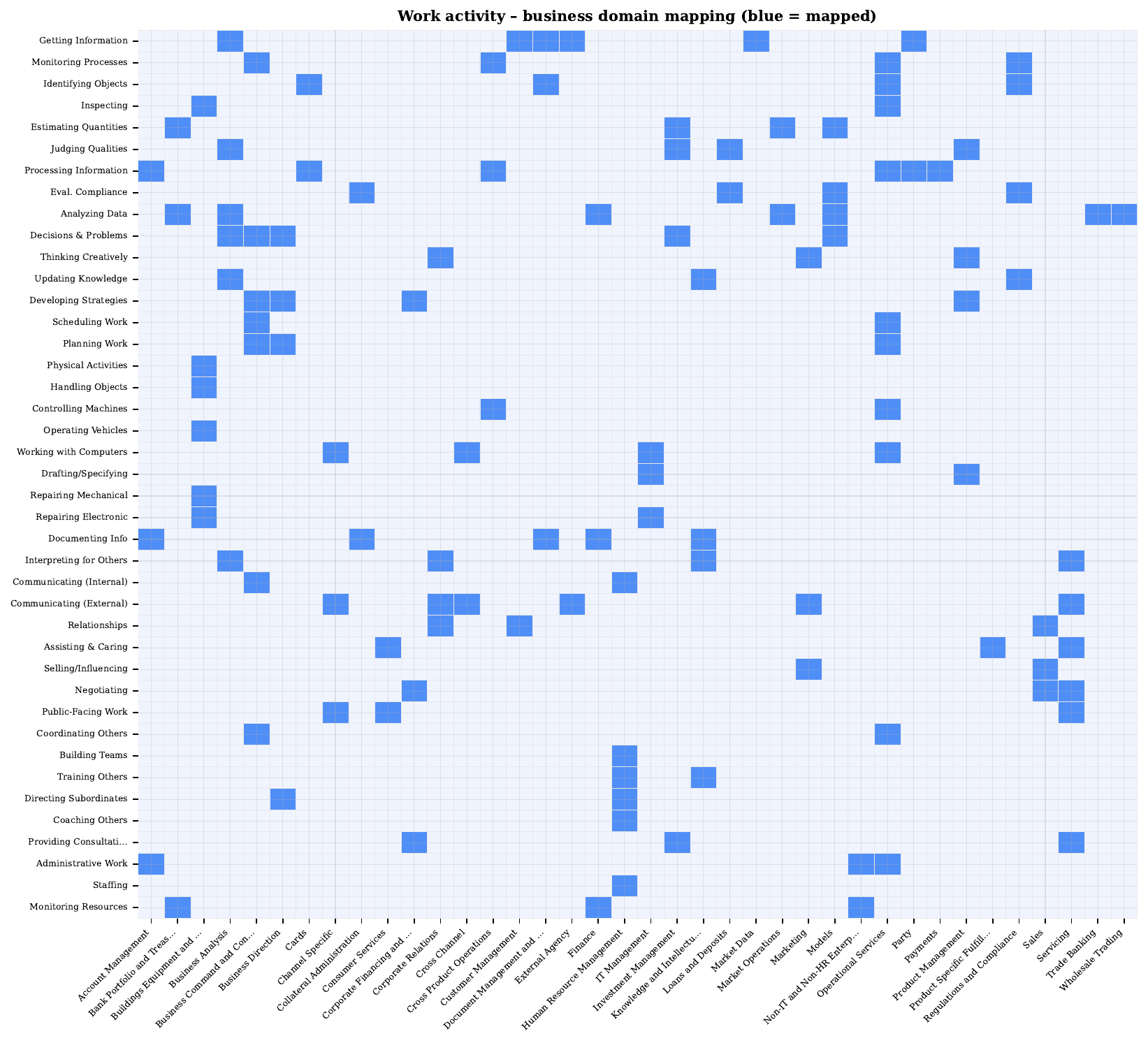}
\caption{Work-activity to business-domain mapping. Blue cells indicate that
benchmark evidence from the work activity is used when computing the
business-domain score.}
\label{fig:capability-heatmap}
\end{figure}

The full 41-GWA × 38-domain mapping is shown as a binary heatmap in
Figure \ref{fig:capability-heatmap}. Each row is a work activity; each
column is a business domain; a blue cell indicates that the work
activity contributes evidence to that domain. The complete
benchmark-to-activity and activity-to-domain tables are provided in
Appendix \ref{sec:appendix}.

Several structural observations follow from the mapping:

\begin{itemize}
\tightlist
\item
  \textbf{Analyzing Data or Information} is the most broadly mapped work
  activity, contributing to seven business domains, reflecting that
  analytical reasoning over data underpins risk, treasury, markets, and
  finance work alike.
\item
  \textbf{Getting Information}, \textbf{Processing Information}, and
  \textbf{Communicating with People Outside the Organization} each
  contribute to six domains, spanning the full spectrum from
  reference-data management to customer-facing service.
\item
  The information-processing GWAs that public benchmarks exercise most
  heavily --- \textbf{Processing Information} (over 300 mapped
  benchmarks) and \textbf{Making Decisions and Solving Problems} (over
  300) --- concentrate their evidence in the Operations and Execution
  and Risk and Compliance areas, while the physical and interpersonal
  GWAs remain unmeasured.
\end{itemize}

\section{Validation}\label{sec:validation}

The framework makes several design choices (the multiplicative K-factor
weighting formula, the pairwise Elo tournament, the
work-activity-to-domain mapping) that could individually or jointly
affect model rankings in ways that are not apparent from the score
itself. This section presents four empirical checks: sensitivity of
rankings to the K-factor weighting formula, a factor ablation study,
rank divergence from global composite rankings, and evidence density
across business domains.

\subsection{Sensitivity to K-Factor Weighting
Scheme}\label{sec:sensitivity}

To assess whether the K-factor weighting formula produces rankings that
are stable across alternative formulations, we compare four K-scaling
schemes for the pairwise Elo tournament:

\begin{enumerate}
  \item \textbf{Uniform K}: each benchmark receives the same base K-factor ($K = 32$).
  \item \textbf{Discrimination K}: $K \propto \text{disc}_b$ (stdev of top-10).
  \item \textbf{Disc.\ $\times$ Coverage K}: $K \propto \text{disc}_b \times
    \text{coverage}_b$, omitting the recency factor.
  \item \textbf{Full K (D$\times$C$\times$R)}: the production formula described in
    Section \ref{sec:weighting}.
\end{enumerate}

For each scheme, we run the 20-pass pairwise Elo tournament and compute
model rankings across four BIAN business domains (IT Management,
Regulations and Compliance, Customer Management, and Market Operations),
each drawn from a distinct Business Area. Spearman rank correlation
\(\rho\) is computed between all pairs, averaged across the four
domains. Spearman \(\rho\) measures how closely two rankings agree,
ranging from 1.0 (identical order) through 0 (no relationship) to
\(-1.0\) (reversed order); a value near 0.90 means two methods place
models in almost the same order.

\begin{figure}[ht]
\centering
\includegraphics[width=\columnwidth]{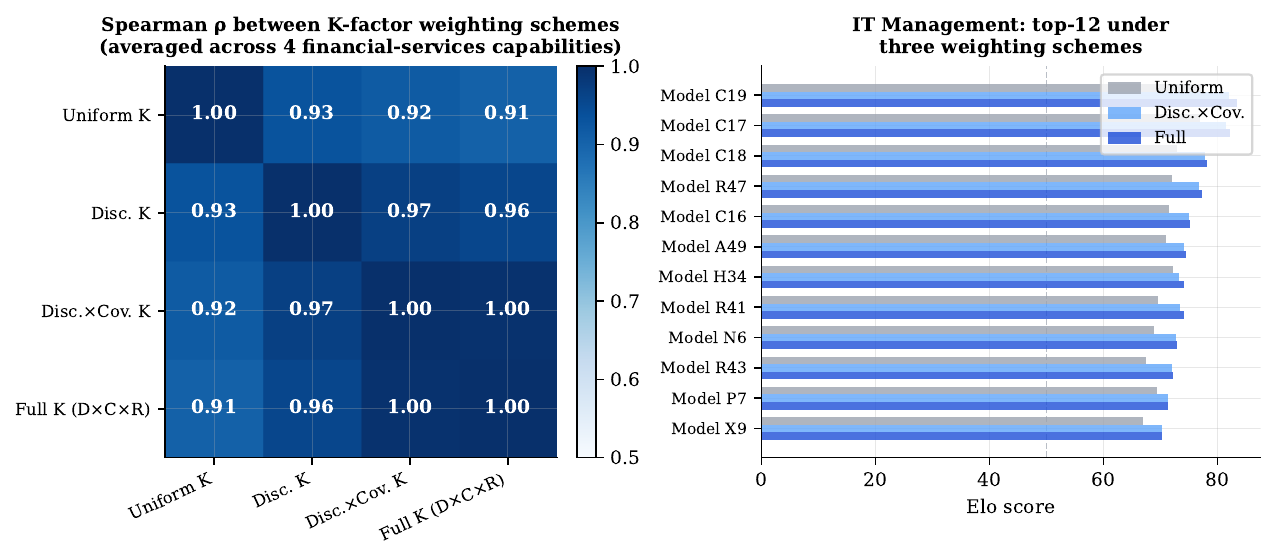}
\caption{Left: Spearman $\rho$ between four K-factor weighting schemes, averaged across
four BIAN business domains. All pairs exceed $\rho = 0.90$. Right: IT
Management Elo scores for the top-12 models under three representative schemes.
Rankings are broadly consistent; the full formula makes modest adjustments
for recent evaluation coverage.}
\label{fig:rank-sensitivity}
\end{figure}

The resulting \(4 \times 4\) correlation matrix (Figure
\ref{fig:rank-sensitivity}, left) shows that rankings are highly stable
across K-factor schemes (\(\rho > 0.90\) in all pairs), with the
smallest correlation between uniform and full K weights
(\(\rho \approx 0.91\)). This suggests that the multiplicative K-factor
formula does not introduce large-scale rank inversions relative to
simpler alternatives; it refines rather than reorders the broad
distribution.

The top-12 IT Management rankings under the uniform,
Disc.\(\times\)Cov., and full K-factor schemes (Figure
\ref{fig:rank-sensitivity}, right) confirm that models at the very top
of the frontier are consistent across schemes; the main effect of adding
recency weighting is to modestly adjust models whose recent evaluation
coverage differs substantially across schemes.

\subsection{Factor Ablation}\label{sec:ablation}

To examine the contribution of individual weight factors, Table
\ref{tab:ablation} shows the top-10 IT Management rankings under
discrimination-only, discrimination\(\times\)coverage, and the full
formula. Ranks are shown as \(\Delta\) from the full-formula ordering.

\begin{table}[ht]
\centering
\small
\caption{IT Management ranking stability under weighting ablations
(top-10 models by full-formula rank). $\Delta$ = rank position change relative
to full-formula ordering. The top three positions are stable across all schemes;
mid-tier models in the top-10 shift by up to 5 positions, suggesting that
discrimination is the dominant factor but that coverage and recency provide
meaningful refinement for models with uneven benchmark participation.}
\label{tab:ablation}
\resizebox{\columnwidth}{!}{\begin{tabular}{p{3.8cm}rrr}
\toprule
Model & Disc.\ only $\Delta$ & Disc.${}\times{}$Cov.\ $\Delta$ & Full rank \\
\midrule
Model C19 & $+1$ & $0$ & 1 \\
Model C17 & $-1$ & $0$ & 2 \\
Model C18 & $0$ & $0$ & 3 \\
Model R47 & $0$ & $0$ & 4 \\
Model C16 & $+1$ & $0$ & 5 \\
Model A49 & $-1$ & $0$ & 6 \\
Model H34 & $+2$ & $+1$ & 7 \\
Model R41 & $+4$ & $-1$ & 8 \\
Model N6 & $-1$ & $0$ & 9 \\
Model R43 & $+1$ & $0$ & 10 \\
\bottomrule
\end{tabular}}
\end{table}

The ablation confirms that discrimination is the dominant factor:
removing coverage and recency does not materially change the top-3
ordering. Adding coverage penalises benchmarks that are rarely
attempted, narrowing the effective evaluation set for specialised work
activities where fewer models are systematically tested. Adding recency
further down-weights benchmarks whose most recent results are from
models released 18--24 months ago, which primarily affects saturated
general-knowledge tests and models that last reported coding results on
older suites.

\subsection{Rank Divergence from Global Composite
Rankings}\label{sec:divergence}

\begin{figure}[ht]
\centering
\includegraphics[width=\columnwidth]{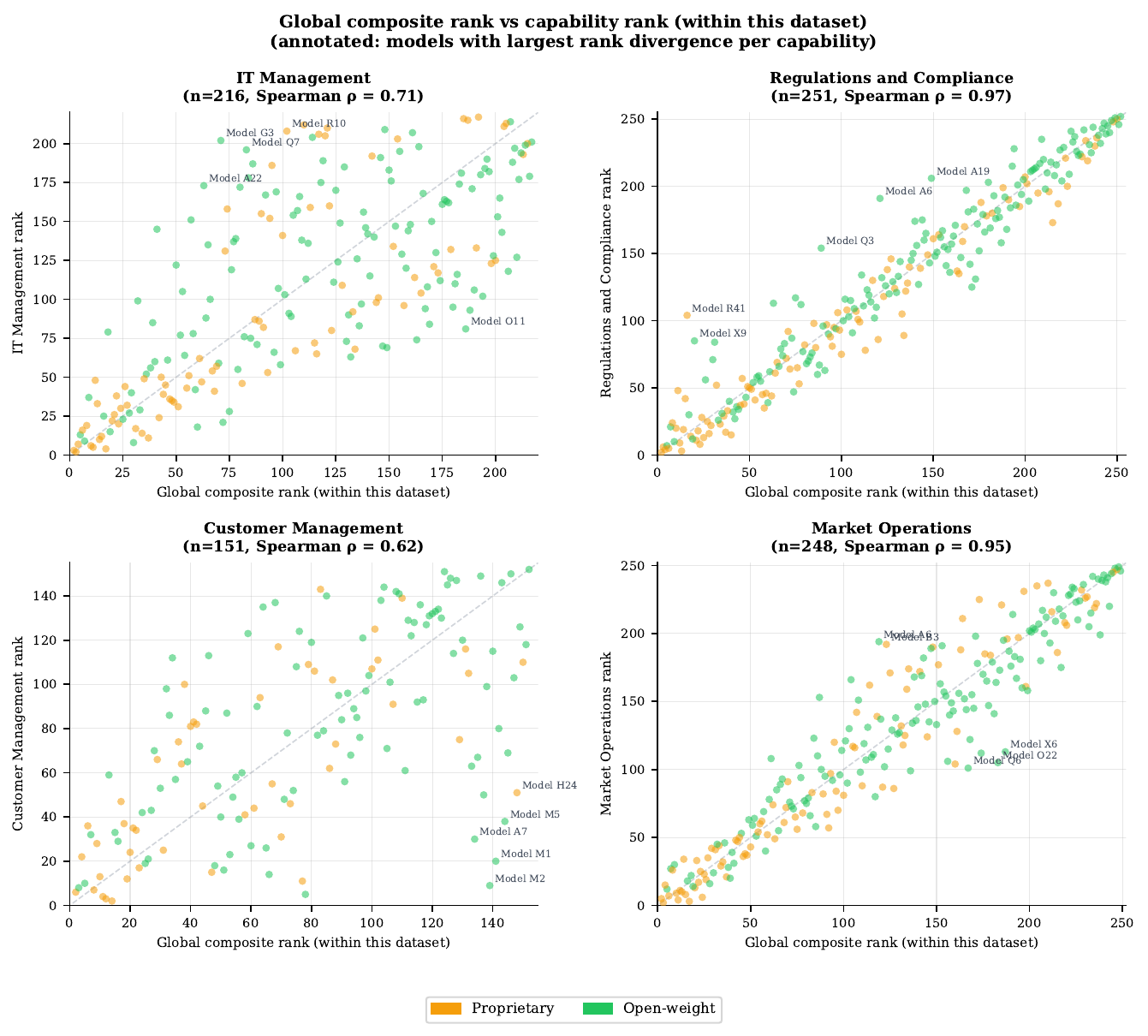}
\caption{Global composite rank vs per-domain rank for four
key BIAN business domains, computed within this dataset (n = models with both
global and domain Elo scores). Amber = proprietary, green = open-weight.
Diagonal line indicates no divergence; annotated models have the five largest
rank differences per panel.}
\label{fig:rank-divergence}
\end{figure}

A key motivation for domain-specific aggregation is that broad composite
rankings do not necessarily predict relative capability for specific
financial-services use cases. We deliberately do not compare each
business domain against a broad financial-services-average score: that
average is itself built largely from the general reasoning, coding, and
factuality activities that drive the global ranking, so the two would
agree almost by construction and tell us nothing. Instead, the global
composite rank (the unweighted mean of all work-activity Elos) is
plotted against four individual business-domain ranks in Figure
\ref{fig:rank-divergence}: IT Management, Customer Management,
Regulations and Compliance, and Market Operations, each computed from
the single-domain Elo.

Per-domain Spearman \(\rho\) values are shown in each panel of Figure
\ref{fig:rank-divergence}. The annotated outliers in each panel
illustrate the practical decision value of the framework: models whose
global composite rank diverges from their ranking on a specific business
domain. The magnitude of divergence varies across domains and across the
pool of models reporting scores in each.

\subsection{Evidence Density by Business
Domain}\label{sec:evidence-density}

Business-domain scores are not equally well-evidenced across all 38
domains. It is important to distinguish two separate concepts:
\emph{catalogue breadth} (the number of benchmark identifiers feeding a
domain, via Figure \ref{fig:benchmark-counts}) and
\emph{model evidence density} (the fraction of models in the dataset
that have actually reported scores on those benchmarks). A domain with
many benchmarks may still have low model evidence density if the
relevant benchmarks are rarely run. For example, a domain whose
contributing benchmarks almost no model has actually run has high
catalogue breadth but low evidence density: many rulers, few
measurements.

\begin{figure}[ht]
\centering
\includegraphics[width=\columnwidth]{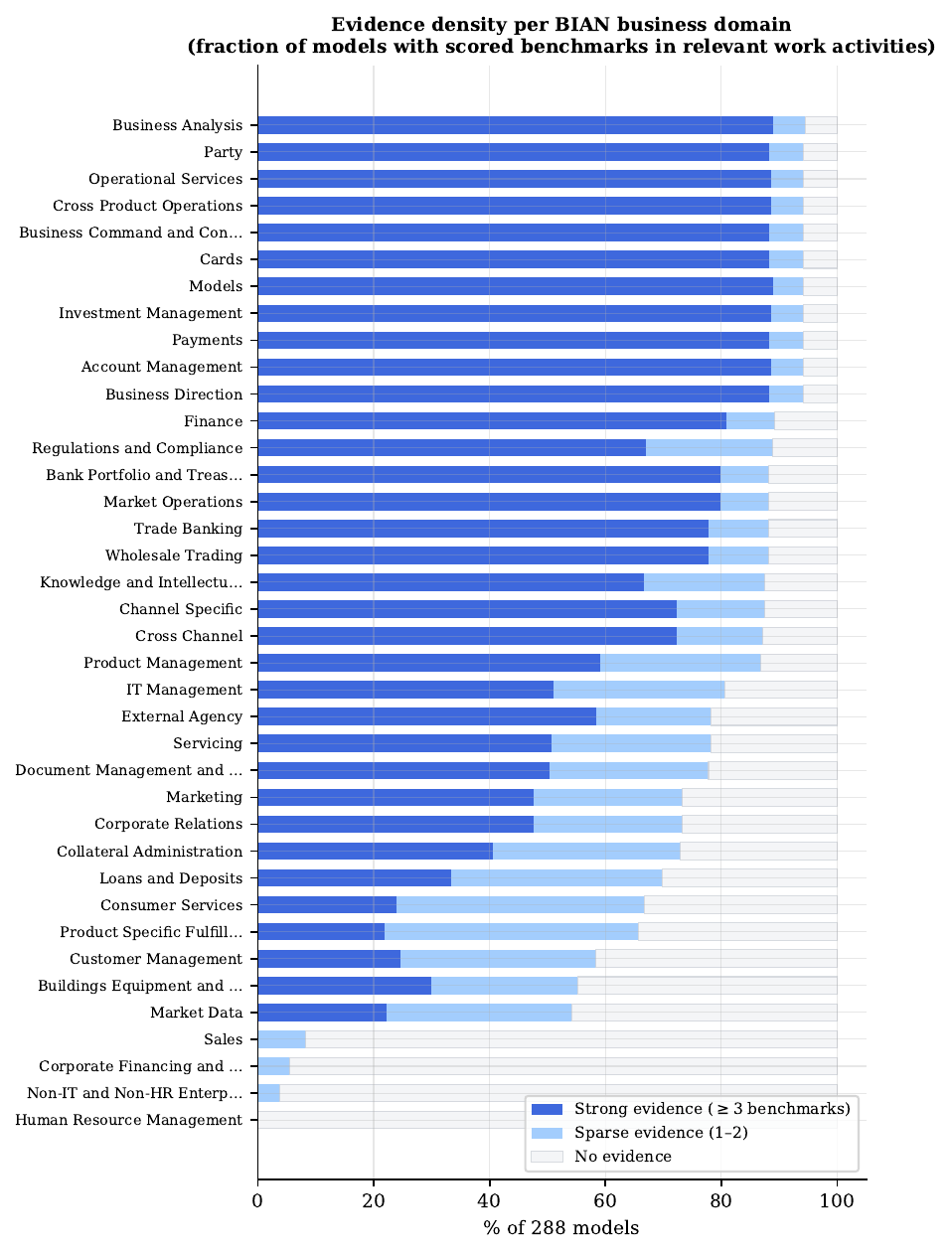}
\caption{Model evidence density per BIAN business domain.
Dark blue = strong evidence ($\geq$3 relevant benchmark scores); light blue =
sparse (1--2); grey = no evidence. Domains drawing on coding and
general-reasoning work activities have the densest model coverage, while
domains reliant on specialist regulatory or financial benchmarks have notably
sparse evidence, reflecting gaps in public benchmark availability for regulated
domain tasks.}
\label{fig:evidence-coverage}
\end{figure}

Model evidence density per business domain, the fraction of models with
strong (\(\geq 3\) relevant benchmark scores), sparse (1--2 benchmarks),
or no evidence, is shown in Figure \ref{fig:evidence-coverage}.

Domains dependent on specialist benchmarks --- those drawing on
Regulations and Compliance, Models, and Finance work --- have notably
lower model evidence density, reflecting a genuine gap in the public
evaluation landscape for regulated domain tasks: few widely-adopted
benchmarks exist for compliance-specific reasoning, financial-crime
investigation, or Australian regulatory knowledge. For these domains,
the Elo should be treated as a weak proxy signal, and internal
evaluation against institution-specific task distributions is
particularly important.

\section{Applications in Model Selection and
Governance}\label{sec:applications}

The capability profiles produced by the meta-benchmarking framework are
intended as practical inputs to model-screening and governance support
in a regulated institution, not solely research artefacts. This section
describes four categories of application, each targeting a distinct risk
or governance need.

\subsection{Model selection}\label{sec:app-selection}

\begin{figure}[ht]
\centering
\includegraphics[width=\columnwidth]{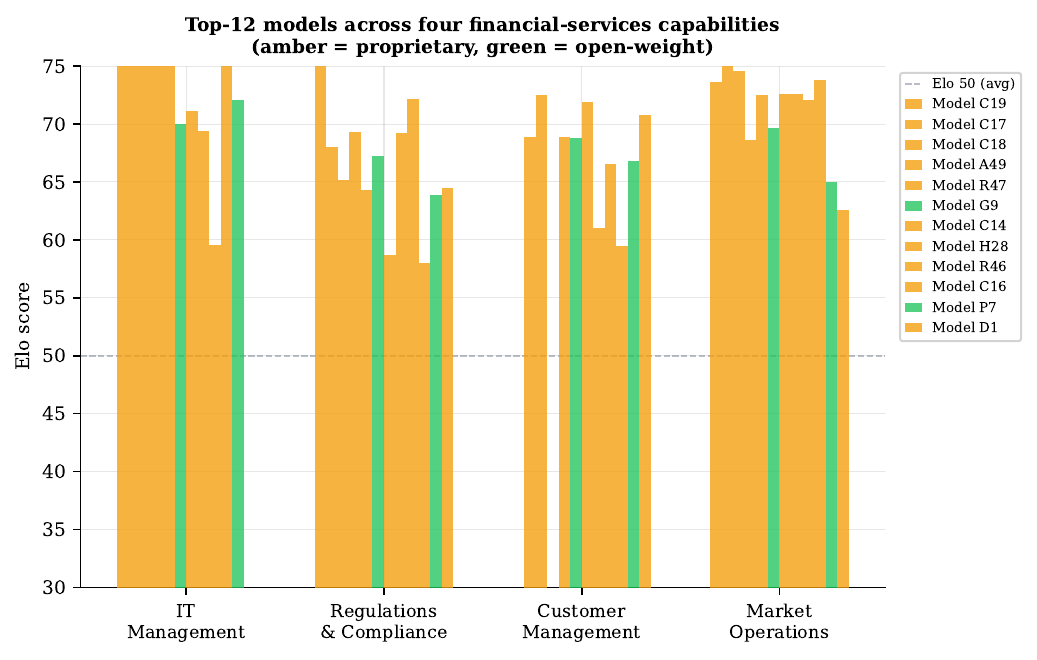}
\caption{Top-12 models ranked by business-domain Elo across four
BIAN business domains. Amber bars indicate proprietary models; green bars
indicate open-weight models. Rankings vary substantially across domains,
and open-weight models are competitive or leading on several domains,
motivating domain-specific rather than global candidate screening.}
\label{fig:capability-comparison}
\end{figure}

The first application is identifying strong candidate models for a use
case before any use-case-specific evaluation is built. Public
leaderboards rank models by global aggregate score; a model that leads
overall may rank poorly on the specific business domains a use case
requires. A use case is mapped to one or more business domains from the
taxonomy (Section \ref{sec:capabilities}), and the domain scores produce
a shortlist of plausible candidates with a principled, traceable basis
grounded in the relevant benchmark evidence. For example, a contract
review use case maps primarily to the Regulations and Compliance and
Document Management and Archive domains; a model that leads on general
reasoning but lacks specialised legal and document handling performance
can be correctly de-prioritised as a candidate despite its global
ranking. The framework narrows the field of models worth evaluating; the
selection decision itself rests on a purpose-built evaluation written
for that use case, able to test the shortlisted candidates against
representative task data. The reasoning behind a candidate shortlist can
be traced through business-domain scores to the underlying benchmark
evidence, making the screening step auditable. Rankings vary
substantially across business domains (Figure
\ref{fig:capability-comparison}), reinforcing that candidates should be
screened against the capabilities a use case actually requires rather
than a global aggregate.

\subsection{Procurement and provider
onboarding}\label{sec:app-procurement}

Procuring a new foundation model or onboarding a new model provider
requires an evidence base that goes beyond provider-supplied benchmark
results, which may be selective or not directly comparable across
providers.

The business-domain profiles provide a provider-agnostic, cross-model
evidence base for procurement evaluation. Candidate models can be
assessed against the same taxonomy regardless of provider, making
like-for-like comparisons tractable. Domain score gaps flag specific
areas that require additional targeted evaluation before onboarding, for
example, if a candidate model scores strongly on IT Management but
poorly on Regulations and Compliance, the procurement team can
commission targeted testing in that gap rather than running a full
bespoke evaluation suite.

This is particularly valuable given the challenges of internal
benchmarking described in Section \ref{sec:internal-challenges}: the
framework provides a credible first-pass filter that substantially
narrows the scope of any subsequent internal evaluation.

\subsection{Concentration risk and fallback
resilience}\label{sec:app-concentration}

Deploying a small number of foundation models across many use cases
creates concentration risk: dependence on a single provider's
availability, pricing decisions, acceptable-use policies, or model
deprecation timelines. For a regulated financial institution, this risk
is material and must be actively monitored.

Capability profiles make concentration risk quantifiable. If several
high-priority use cases all require a specific capability for which only
one provider's model scores above the acceptable threshold, that
dependency is visible and can be surfaced in risk reporting. The same
profiles support systematic fallback specification: for each deployed
model, candidate substitutes are the next-highest-scoring models on the
relevant capability from a \emph{different} provider, within an
acceptable Elo delta. The delta quantifies the performance cost of
failover: a \(-2\) Elo fallback is materially different from a \(-10\)
Elo fallback, and this cost can be factored into operational risk
tolerances. This matters most during availability incidents (an API
outage, rate limit breach, or sudden policy change that interrupts
access to a deployed model). Any fallback arrangement would require
separate operational, security, legal and risk review before
implementation.

\subsection{Ongoing governance and frontier
tracking}\label{sec:app-governance}

Model performance is not static. As new models are released, the
relative capability standing of currently deployed models changes. A
model that ranked highly in the capability profiles twelve months ago
may now be surpassed by a margin that warrants re-evaluation.

The framework produces updated capability profiles as new benchmark data
becomes available, enabling institutions to track capability drift over
time. Governance policies can be expressed in quantitative terms, for
instance by defining a threshold delta above which new public evidence
warrants re-evaluation of a currently deployed model. This converts
ongoing model governance from an ad-hoc activity into a measurable,
threshold-driven process, in which committees review structured
capability drift reports rather than interpreting raw benchmark outputs.

\section{Assumptions and Limitations}\label{sec:limitations}

We describe the key assumptions underlying the framework and the
limitations that follow from them.

\textbf{Rolling window definition.} The rolling window (default: 12
months) defines ``recent'' for both discrimination and coverage
calculations. A shorter window is more responsive to newly released
models but is noisier: a benchmark with only 3 recent models receives
the minimum qualifying score. A longer window is more stable but may
include models that are no longer competitive at the frontier. The
window is configurable; the default of 12 months represents a judgement
that models released more than a year ago are unlikely to represent
current frontier capability.

\textbf{Missing scores treated as neutral.} Models absent from a
benchmark contribute no pairwise comparisons for that benchmark. Their
Elo is determined entirely by benchmarks they did attempt. This avoids
the distortion of imputing \(z = 0\) (benchmark mean) but introduces a
cold-start effect: a model with very few benchmark results participates
in few comparisons and its rating is determined by a small sample. Such
models are flagged as low-evidence in the capability evidence density
analysis (Section \ref{sec:validation}).

\textbf{Pairwise Elo scores shift with new data.} Because Elo is
recomputed from scratch on every pipeline run, scores are not stable
across snapshots. Adding a new frontier model compresses the ratings of
existing models; adding new benchmarks dilutes the impact of existing
ones. Score-shift is bounded by the K-factor and typical changes between
runs are small (\(< 2\) Elo points for established models), but users
should not compare absolute scores across runs that span major data
additions. The reference point of 50 is likewise relative: it is the
mean rating of all models in the tournament for a given task after
normalisation, not a fixed reference model, and its meaning shifts as
the model population changes. Convergence is also approximate rather
than guaranteed: twenty passes are sufficient for stable ranks at the
model and benchmark counts in this dataset (rank changes between pass 15
and pass 20 are \(< 1\%\) of model-task pairs, and sensitivity to
K-factor variants is low; Section \ref{sec:validation}), but the
procedure is not formally proven to reach a fixed point. The Elo display
scale is an interpretability aid and does not claim a formal
relationship with the rating system used in competitive games.

\textbf{Equal-weight, independent capability averaging.} The default
equal-weight aggregation means a consistently mediocre model can score
similarly to one that excels on one task and fails on others (both near
Elo 50). It also treats task scores as independent inputs, whereas model
capabilities are in practice correlated: a model that reasons well
generally also tends to perform well on domain-specific reasoning. The
aggregation does not model this correlation structure and may therefore
over- or under-represent the benefit of task specialisation. This is
deliberate: the use-case evaluator is expected to examine individual
task scores, and an implementation may allow task weights to be adjusted
for specific priorities.

\textbf{Self-reported scores.} All benchmark scores are sourced from
publicly reported results. Self-reported scores (published by model
providers in technical reports) and independently reproduced scores are
mixed in the dataset. No independent verification of reported scores is
performed. Scores on benchmarks known to be present in pre-training
corpora may be overstated, a risk that applies especially to older,
widely known benchmarks (Xu et al. 2024). The dynamic weighting
partially mitigates this: if all frontier models score near-perfectly on
a contaminated benchmark, its discrimination score falls to near zero
and it is effectively excluded from recent task score computation.

\textbf{LLM Stats coverage gaps.} The upstream data source may not index
all recently released models; some organisations publish technical
reports substantially after model release, causing score availability to
lag model availability. The pipeline skips models with no scores rather
than imputing values. Models released less than approximately two weeks
before the last data refresh may not appear in the dataset.

\textbf{Manual taxonomy.} The benchmark-to-activity and
activity-to-domain mappings are maintained manually in structured
configuration files. The taxonomy grounds the activity layer in the
O*NET Generalized Work Activities catalogue and the domain layer in the
BIAN Service Landscape, but the assignment of benchmarks to activities
and activities to domains is subject to the judgement of the maintainer.
As new use-case patterns emerge, particularly in multimodal, agentic,
and voice-based interaction, the mappings require review and extension.
Of the 41 work activities, 17 (predominantly physical and managerial
activities such as operating vehicles, staffing, and resolving
conflicts) have no public benchmark coverage and therefore contribute no
evidence to the domains they map to.

\textbf{Non-random missingness.} Models that are heavily evaluated tend
to be frontier proprietary systems from organisations with large
evaluation budgets or active research communities. Smaller open-weight
models and non-English models are underrepresented in public benchmark
coverage. This structural bias means capability scores are most reliable
for the most-evaluated models and carry greater uncertainty for
less-reported models, a distinction not surfaced by the score alone.

\textbf{Thin finance and compliance coverage; regulatory obligations not
benchmarked.} Public benchmarks for general reasoning and coding are
abundant and frequently updated, whereas those targeting financial
question answering, compliance reasoning, and AML/KYC tasks are
relatively sparse (Islam et al. 2023; Q. Xie et al. 2024). Business
domains such as Regulations and Compliance, Models, and Finance
therefore rely more heavily on proxy work activities (e.g.~general
reasoning, getting and evaluating information) and carry higher
uncertainty than technology-focused domains. Crucially, the scoring does
not directly test compliance with jurisdiction-specific obligations: a
model that scores well on regulatory and compliance benchmarks may still
fail institution-specific policy controls or respond incorrectly to
local regulatory definitions. APRA's 2026 guidance (Australian
Prudential Regulation Authority 2026) identifies accountability,
auditability, and human oversight as governance requirements that
benchmark scores do not assess.

\textbf{Composite scores can mask critical failures.} A high average
domain score can coexist with an unacceptable score on a risk-critical
work activity (e.g.~a high Customer Management Elo that masks poor
performance on activities requiring factual grounding). Composite scores
should always be examined alongside individual work-activity scores for
safety-sensitive domains, and threshold or red-flag checks should be
applied before deployment (National Institute of Standards and
Technology 2024; Guldimann et al. 2024). More broadly, as described in
Section \ref{sec:capabilities}, a business-domain score is a screening
tool, not a deployment certification: production use in a regulated
institution requires additional technical, legal, and governance review
outside the scope of this framework (Bank for International Settlements
Financial Stability Institute 2024; Australian Securities and
Investments Commission 2024).

\textbf{Ordinal-only pairwise comparisons.} The pairwise Elo tournament
uses win/draw/loss outcomes rather than score margins. A model that wins
a benchmark by 0.1 percentage points and one that wins by 20 percentage
points both receive a win. This discards margin information and may be
sensitive to benchmark noise or rounding near boundary score values. A
draw threshold, treating near-ties as draws when scores are within a
small epsilon of one another, would partially address this, but the
appropriate threshold is benchmark-specific and is not currently
implemented.

\section{Discussion and Future Work}\label{sec:conclusion}

We have described a meta-benchmarking framework that maps 452 publicly
reported benchmarks into 41 O*NET work activities and 38 BIAN banking
business domains, covering 288 models from 25 organisations. Its central
contribution is a dynamic weighting scheme (discrimination × coverage ×
recency) that automatically promotes benchmarks actively contesting
frontier performance and suppresses saturated legacy tests, without
manual score curation or benchmark lifecycle management. Sensitivity
analysis, factor ablation, and rank-divergence checks indicate the
resulting rankings are stable and consistent with independent evidence.

The framework addresses a practical gap: public leaderboards optimise
for breadth and global average performance, while regulated institutions
need evidence about specific business domains. By grounding the taxonomy
in established external standards (O*NET work activities and the BIAN
Service Landscape) rather than academic convention, the business-domain
scores give institutions an auditable, provider-agnostic basis for
identifying candidate models, scoping procurement evaluations,
quantifying concentration risk, and tracking capability drift over time,
narrowing the field before any use-case-specific evaluation is built
rather than replacing it.

Rather than introduce yet another bespoke capability framework, we
aligned the two abstraction layers to standards that already exist,
O*NET Generalized Work Activities for the task layer and the BIAN
Service Landscape for the domain layer. This enables future validation
because the layers are defined by external authorities, the mappings can
be checked against a reference outside the paper, and independent
parties can re-map the same benchmarks and compare. Second, it enables
extensibility, extends into O*NET's other task, skill and occupation
dimensions (including Intermediate and Detailed Work Activities, and the
underlying skills and abilities), while the domain layer can be refined
into the full BIAN service-domain hierarchy.

The benchmark-to-activity and activity-to-domain mappings are a
candidate mapping but not a validated ground truth. The alignment to
external standards allows further scrutiny and validation based on
ground truth data, and improved as the benchmark ecosystem evolves.

Finally, the framework does not yet account for the cost of task
completion. A model that reaches a given domain score at a fraction of
the inference cost may be the better choice in practice, and future work
could fold price and latency into the weighting so that scores reflect
not just capability but the efficiency with which it is delivered.

\appendix

\section{Taxonomy Tables}\label{sec:appendix}

This appendix provides the complete benchmark-to-work-activity and
work-activity-to-domain mappings in tabular form. All data reflects the
state of the taxonomy configuration files as of June 2026.

\subsection{Work-Activity-to-Domain
Mapping}\label{work-activity-to-domain-mapping}

Table \ref{tab:task-cap} shows the 38 BIAN business domains and the
O*NET work activities that contribute evidence to each.

\onecolumn

\begin{longtable}{p{3.6cm}p{10.7cm}}
\caption{BIAN business-domain to work-activity mapping. Each domain lists the O*NET Generalized Work Activities mapped to it.}
\label{tab:task-cap} \\
\toprule
\textbf{Business Domain} & \textbf{Contributing Work Activities} \\
\midrule
\endfirsthead
\toprule
\textbf{Business Domain} & \textbf{Contributing Work Activities} \\
\midrule
\endhead
Account Management & Processing Information, Documenting/Recording Information, Performing Administrative Activities \\
\midrule
Bank Portfolio and Treasury & Estimating the Quantifiable Characteristics of Products, Events, or Information, Analyzing Data or Information, Monitoring and Controlling Resources \\
\midrule
Buildings Equipment and Facilities & Inspecting Equipment, Structures, or Materials, Performing General Physical Activities, Handling and Moving Objects, Operating Vehicles, Mechanized Devices, or Equipment, Repairing and Maintaining Mechanical Equipment, Repairing and Maintaining Electronic Equipment \\
\midrule
Business Analysis & Getting Information, Judging the Qualities of Objects, Services, or People, Analyzing Data or Information, Making Decisions and Solving Problems, Updating and Using Relevant Knowledge, Interpreting the Meaning of Information for Others \\
\midrule
Business Command and Control & Monitoring Processes, Materials, or Surroundings, Making Decisions and Solving Problems, Developing Objectives and Strategies, Scheduling Work and Activities, Organizing, Planning, and Prioritizing Work, Communicating with Supervisors, Peers, or Subordinates, Coordinating the Work and Activities of Others \\
\midrule
Business Direction & Making Decisions and Solving Problems, Developing Objectives and Strategies, Organizing, Planning, and Prioritizing Work, Guiding, Directing, and Motivating Subordinates \\
\midrule
Cards & Identifying Objects, Actions, and Events, Processing Information \\
\midrule
Channel Specific & Working with Computers, Communicating with People Outside the Organization, Performing for or Working Directly with the Public \\
\midrule
Collateral Administration & Evaluating Information to Determine Compliance with Standards, Documenting/Recording Information \\
\midrule
Consumer Services & Assisting and Caring for Others, Performing for or Working Directly with the Public \\
\midrule
Corporate Financing and Advisory Services & Developing Objectives and Strategies, Resolving Conflicts and Negotiating with Others, Providing Consultation and Advice to Others \\
\midrule
Corporate Relations & Thinking Creatively, Interpreting the Meaning of Information for Others, Communicating with People Outside the Organization, Establishing and Maintaining Interpersonal Relationships \\
\midrule
Cross Channel & Working with Computers, Communicating with People Outside the Organization \\
\midrule
Cross Product Operations & Monitoring Processes, Materials, or Surroundings, Processing Information, Controlling Machines and Processes \\
\midrule
Customer Management & Getting Information, Establishing and Maintaining Interpersonal Relationships \\
\midrule
Document Management and Archive & Getting Information, Identifying Objects, Actions, and Events, Documenting/Recording Information \\
\midrule
External Agency & Getting Information, Communicating with People Outside the Organization \\
\midrule
Finance & Analyzing Data or Information, Documenting/Recording Information, Monitoring and Controlling Resources \\
\midrule
Human Resource Management & Communicating with Supervisors, Peers, or Subordinates, Developing and Building Teams, Training and Teaching Others, Guiding, Directing, and Motivating Subordinates, Coaching and Developing Others, Staffing Organizational Units \\
\midrule
IT Management & Working with Computers, Drafting, Laying Out, and Specifying Technical Devices, Parts, and Equipment, Repairing and Maintaining Electronic Equipment \\
\midrule
Investment Management & Estimating the Quantifiable Characteristics of Products, Events, or Information, Judging the Qualities of Objects, Services, or People, Making Decisions and Solving Problems, Providing Consultation and Advice to Others \\
\midrule
Knowledge and Intellectual Property Management & Updating and Using Relevant Knowledge, Documenting/Recording Information, Interpreting the Meaning of Information for Others, Training and Teaching Others \\
\midrule
Loans and Deposits & Judging the Qualities of Objects, Services, or People, Evaluating Information to Determine Compliance with Standards \\
\midrule
Market Data & Getting Information \\
\midrule
Market Operations & Estimating the Quantifiable Characteristics of Products, Events, or Information, Analyzing Data or Information \\
\midrule
Marketing & Thinking Creatively, Communicating with People Outside the Organization, Selling or Influencing Others \\
\midrule
Models & Estimating the Quantifiable Characteristics of Products, Events, or Information, Evaluating Information to Determine Compliance with Standards, Analyzing Data or Information, Making Decisions and Solving Problems \\
\midrule
Non-IT and Non-HR Enterprise Services & Performing Administrative Activities, Monitoring and Controlling Resources \\
\midrule
Operational Services & Monitoring Processes, Materials, or Surroundings, Identifying Objects, Actions, and Events, Inspecting Equipment, Structures, or Materials, Processing Information, Scheduling Work and Activities, Organizing, Planning, and Prioritizing Work, Controlling Machines and Processes, Working with Computers, Coordinating the Work and Activities of Others, Performing Administrative Activities \\
\midrule
Party & Getting Information, Processing Information \\
\midrule
Payments & Processing Information \\
\midrule
Product Management & Judging the Qualities of Objects, Services, or People, Thinking Creatively, Developing Objectives and Strategies, Drafting, Laying Out, and Specifying Technical Devices, Parts, and Equipment \\
\midrule
Product Specific Fulfillment & Assisting and Caring for Others \\
\midrule
Regulations and Compliance & Monitoring Processes, Materials, or Surroundings, Identifying Objects, Actions, and Events, Evaluating Information to Determine Compliance with Standards, Updating and Using Relevant Knowledge \\
\midrule
Sales & Establishing and Maintaining Interpersonal Relationships, Selling or Influencing Others, Resolving Conflicts and Negotiating with Others \\
\midrule
Servicing & Interpreting the Meaning of Information for Others, Communicating with People Outside the Organization, Assisting and Caring for Others, Resolving Conflicts and Negotiating with Others, Performing for or Working Directly with the Public, Providing Consultation and Advice to Others \\
\midrule
Trade Banking & Analyzing Data or Information \\
\midrule
Wholesale Trading & Analyzing Data or Information \\
\bottomrule
\end{longtable}

\subsection{Full Benchmark-to-Work-Activity
Mapping}\label{sec:full-table}

Table \ref{tab:full-bench-task} lists each O*NET work activity with
public benchmark coverage, the count of benchmarks assigned to each, and
the full set of benchmark identifiers. Benchmark identifiers correspond
to the LLM Stats API slug format (lowercase, hyphenated). Many
benchmarks appear in multiple work activities; the total count of
benchmark--work-activity assignments is 1\{,\}750 across 452 unique
benchmark identifiers, spanning 24 of the 41 work activities (the
remaining 17 are physical and managerial activities with no public
benchmark coverage).

\begin{longtable}{p{4.2cm}cp{11.1cm}}
\caption{Full benchmark-to-work-activity mapping (as of June 2026). Each task is an O*NET Generalized Work Activity; benchmark counts reflect unique identifiers assigned to each work activity; a benchmark appearing in multiple activities is counted once per activity.}
\label{tab:full-bench-task} \\
\toprule
\textbf{Work Activity} & \textbf{N} & \textbf{Benchmarks} \\
\midrule
\endfirsthead
\multicolumn{3}{l}{\small\textit{(continued)}} \\
\toprule
\textbf{Work Activity} & \textbf{N} & \textbf{Benchmarks} \\
\midrule
\endhead
\midrule
\multicolumn{3}{r}{\small\textit{(continued on next page)}} \\
\endfoot
\bottomrule
\endlastfoot
Analyzing Data or Information & 87 & \small{\texttt{acebench}, \texttt{agieval}, \texttt{aime}, \texttt{aime-2024}, \texttt{aime-2025}, \texttt{aime-2026}, \texttt{alignbench}, \texttt{amc-2022-23}, \texttt{bbh}, \texttt{beyond-aime}, \texttt{bfcl-v3}, \texttt{big-bench}, \texttt{big-bench-hard}, \texttt{bixbench}, \texttt{cbnsl}, \texttt{cnmo-2024}, \texttt{codeforces}, \texttt{crag}, \texttt{drop}, \texttt{dynamath}, \texttt{finance-agent}, \texttt{finqa}, \texttt{finsearchcomp-t2-t3}, \texttt{finsearchcomp-t3}, \texttt{french-mmlu}, \texttt{frontiermath}, \texttt{frontierscience-research}, \texttt{functionalmath}, \texttt{gdpval-aa}, \texttt{gdpval-mm}, \texttt{genebench}, \texttt{global-piqa}, \texttt{gpqa}, \texttt{gsm-8k-(cot)}, \texttt{gsm8k}, \texttt{gsm8k-chat}, \texttt{hiddenmath}, \texttt{hmmt-2025}, \texttt{hmmt-feb-26}, \texttt{hmmt25}, \texttt{humanity's-last-exam}, \texttt{imo-answerbench}, \texttt{intergps}, \texttt{ipho-2025}, \texttt{livebench}, \texttt{livebench-20241125}, \texttt{math}, \texttt{math-(cot)}, \texttt{math-500}, \texttt{matharena-apex}, \texttt{mathverse-mini}, \texttt{mathvision}, \texttt{mathvista}, \texttt{mathvista-mini}, \texttt{mgsm}, \texttt{mmlu}, \texttt{mmlu-(cot)}, \texttt{mmlu-base}, \texttt{mmlu-chat}, \texttt{mmlu-french}, \texttt{mmlu-pro}, \texttt{mmlu-prox}, \texttt{mmlu-redux}, \texttt{mmlu-stem}, \texttt{mmmlu}, \texttt{mmvet}, \texttt{mmvetgpt4turbo}, \texttt{multilingual-mgsm-(cot)}, \texttt{olympiadbench}, \texttt{omnimath}, \texttt{openai-mmlu}, \texttt{perceptiontest}, \texttt{phibench}, \texttt{physicsfinals}, \texttt{piqa}, \texttt{polymath}, \texttt{polymath-en}, \texttt{sat-math}, \texttt{scienceqa}, \texttt{stem}, \texttt{supergpqa}, \texttt{theoremqa}, \texttt{truthfulqa}, \texttt{usamo25}, \texttt{we-math}, \texttt{wmdp}, \texttt{writingbench}} \\
\midrule
Assisting and Caring for Others & 21 & \small{\texttt{acebench}, \texttt{french-mmlu}, \texttt{healthbench}, \texttt{healthbench-hard}, \texttt{mmlu}, \texttt{mmlu-(cot)}, \texttt{mmlu-base}, \texttt{mmlu-chat}, \texttt{mmlu-french}, \texttt{mmlu-pro}, \texttt{mmlu-prox}, \texttt{mmmu}, \texttt{mmmu-(val)}, \texttt{mmmu-(validation)}, \texttt{mmmuval}, \texttt{openai-mmlu}, \texttt{supergpqa}, \texttt{truthfulqa}, \texttt{videommmu}, \texttt{wmdp}, \texttt{wmt23}} \\
\midrule
Communicating with People Outside the Organization & 72 & \small{\texttt{alignbench}, \texttt{bbh}, \texttt{big-bench}, \texttt{big-bench-extra-hard}, \texttt{big-bench-hard}, \texttt{boolq}, \texttt{c-eval}, \texttt{charadessta}, \texttt{cluewsc}, \texttt{cmmlu}, \texttt{collie}, \texttt{common-voice-15}, \texttt{commonsenseqa}, \texttt{covost2}, \texttt{covost2-en-zh}, \texttt{csimpleqa}, \texttt{eclektic}, \texttt{fleurs}, \texttt{french-mmlu}, \texttt{global-mmlu}, \texttt{global-mmlu-lite}, \texttt{include}, \texttt{lingoqa}, \texttt{maxife}, \texttt{mega-mlqa}, \texttt{mega-tydi-qa}, \texttt{mega-udpos}, \texttt{mega-xcopa}, \texttt{mega-xstorycloze}, \texttt{mm-mt-bench}, \texttt{mmlu}, \texttt{mmlu-(cot)}, \texttt{mmlu-base}, \texttt{mmlu-chat}, \texttt{mmlu-french}, \texttt{mmlu-pro}, \texttt{mmlu-prox}, \texttt{mmlu-redux}, \texttt{mmmlu}, \texttt{mt-bench}, \texttt{multi-if}, \texttt{multichallenge}, \texttt{multilf}, \texttt{multilingual-mmlu}, \texttt{multipl-e}, \texttt{multipl-e-humaneval}, \texttt{nova-63}, \texttt{open-rewrite}, \texttt{spider}, \texttt{squality}, \texttt{superglue}, \texttt{tau-bench-airline}, \texttt{tau-bench-retail}, \texttt{tau2-airline}, \texttt{tau2-retail}, \texttt{tau2-telecom}, \texttt{tau3-telecom}, \texttt{tldr9+-(test)}, \texttt{translation-en→set1-comet22}, \texttt{translation-en→set1-spbleu}, \texttt{translation-set1→en-comet22}, \texttt{translation-set1→en-spbleu}, \texttt{tydiqa}, \texttt{vatex}, \texttt{voicebench-avg}, \texttt{vqav2-(val)}, \texttt{wild-bench}, \texttt{winogrande}, \texttt{wmt23}, \texttt{wmt24++}, \texttt{writingbench}, \texttt{xlsum-english}} \\
\midrule
Controlling Machines and Processes & 1 & \small{\texttt{robospatialhome}} \\
\midrule
Developing Objectives and Strategies & 2 & \small{\texttt{gdpval-aa}, \texttt{uniform-bar-exam}} \\
\midrule
Documenting/Recording Information & 38 & \small{\texttt{alignbench}, \texttt{alpacaeval-2.0}, \texttt{arena-hard}, \texttt{arena-hard-v2}, \texttt{bfcl-v3}, \texttt{cc-ocr}, \texttt{charxiv-d}, \texttt{collie}, \texttt{complexfuncbench}, \texttt{creative-writing-v3}, \texttt{docvqa}, \texttt{eq-bench}, \texttt{govreport}, \texttt{if}, \texttt{ifeval}, \texttt{internal-api-instruction-following-(hard)}, \texttt{longbench-v2}, \texttt{mm-if-eval}, \texttt{multi-if}, \texttt{ocrbench}, \texttt{ocrbench-v2}, \texttt{ocrbench-v2-(en)}, \texttt{ocrbench-v2-(zh)}, \texttt{omnidocbench-1.5}, \texttt{open-rewrite}, \texttt{qmsum}, \texttt{sifo}, \texttt{sifo-multiturn}, \texttt{simplevqa}, \texttt{squality}, \texttt{summscreenfd}, \texttt{textvqa}, \texttt{tldr9+-(test)}, \texttt{vqav2}, \texttt{vqav2-(test)}, \texttt{vqav2-(val)}, \texttt{writingbench}, \texttt{xlsum-english}} \\
\midrule
Drafting, Laying Out, and Specifying Technical Devices, Parts, and Equipment & 91 & \small{\texttt{aider}, \texttt{aider-polyglot}, \texttt{aider-polyglot-edit}, \texttt{bigcodebench}, \texttt{bigcodebench-full}, \texttt{bigcodebench-hard}, \texttt{bird-sql-(dev)}, \texttt{cc-bench-v2-backend}, \texttt{cc-bench-v2-frontend}, \texttt{cc-bench-v2-repo}, \texttt{cfeval}, \texttt{claw-eval}, \texttt{codegolf-v2.2}, \texttt{crux-o}, \texttt{cruxeval-input-cot}, \texttt{cruxeval-o}, \texttt{cruxeval-output-cot}, \texttt{cybench}, \texttt{cybergym}, \texttt{design2code}, \texttt{evalplus}, \texttt{flame-vlm-code}, \texttt{fullstackbench-en}, \texttt{fullstackbench-zh}, \texttt{gorilla-benchmark-api-bench}, \texttt{humaneval}, \texttt{humaneval+}, \texttt{humaneval-average}, \texttt{humaneval-er}, \texttt{humaneval-mul}, \texttt{humaneval-plus}, \texttt{humanevalfim-average}, \texttt{instruct-humaneval}, \texttt{lbpp-(v2)}, \texttt{livecodebench}, \texttt{livecodebench-pro}, \texttt{livecodebench-v5}, \texttt{livecodebench-v5-24.12-25.2}, \texttt{livecodebench-v6}, \texttt{mbpp}, \texttt{mbpp+}, \texttt{mbpp-++-base-version}, \texttt{mbpp-evalplus}, \texttt{mbpp-evalplus-(base)}, \texttt{mbpp-pass@1}, \texttt{mbpp-plus}, \texttt{mcp-atlas}, \texttt{mle-bench-lite}, \texttt{mm-clawbench}, \texttt{mm-mind2web}, \texttt{mobileminiwob++-sr}, \texttt{multi-swe-bench}, \texttt{multipl-e-mbpp}, \texttt{natural2code}, \texttt{nl2repo}, \texttt{octocodingbench}, \texttt{ojbench}, \texttt{ojbench-cpp}, \texttt{openrca}, \texttt{paperbench}, \texttt{pinchbench}, \texttt{qwenwebbench}, \texttt{repobench}, \texttt{repoqa}, \texttt{seccodebench}, \texttt{skillsbench}, \texttt{swe-bench-multilingual}, \texttt{swe-bench-multimodal}, \texttt{swe-bench-pro}, \texttt{swe-bench-verified}, \texttt{swe-bench-verified-(agentic-coding)}, \texttt{swe-bench-verified-(agentless)}, \texttt{swe-bench-verified-(multiple-attempts)}, \texttt{swe-lancer}, \texttt{swe-lancer-(ic-diamond-subset)}, \texttt{swe-perf}, \texttt{swe-review}, \texttt{swt-bench}, \texttt{terminal-bench}, \texttt{terminal-bench-2}, \texttt{terminus}, \texttt{vibe}, \texttt{vibe-android}, \texttt{vibe-backend}, \texttt{vibe-ios}, \texttt{vibe-pro}, \texttt{vibe-simulation}, \texttt{vibe-web}, \texttt{vision2web}, \texttt{visualwebbench}, \texttt{zclawbench}} \\
\midrule
Establishing and Maintaining Interpersonal Relationships & 6 & \small{\texttt{alignbench}, \texttt{eq-bench}, \texttt{meld}, \texttt{mt-bench}, \texttt{openai-mmlu}, \texttt{social-iqa}} \\
\midrule
Estimating the Quantifiable Characteristics of Products, Events, or Information & 98 & \small{\texttt{agieval}, \texttt{aime}, \texttt{aime-2024}, \texttt{aime-2025}, \texttt{aime-2026}, \texttt{alignbench}, \texttt{amc-2022-23}, \texttt{arc-agi}, \texttt{arc-agi-v2}, \texttt{arkitscenes}, \texttt{bbh}, \texttt{beyond-aime}, \texttt{big-bench}, \texttt{big-bench-hard}, \texttt{blink}, \texttt{cbnsl}, \texttt{cnmo-2024}, \texttt{codeforces}, \texttt{countbench}, \texttt{crag}, \texttt{drop}, \texttt{dynamath}, \texttt{embspatialbench}, \texttt{erqa}, \texttt{finqa}, \texttt{finsearchcomp-t2-t3}, \texttt{finsearchcomp-t3}, \texttt{frontiermath}, \texttt{functionalmath}, \texttt{global-piqa}, \texttt{gpqa}, \texttt{graphwalks-bfs-<128k}, \texttt{graphwalks-bfs->128k}, \texttt{graphwalks-parents-<128k}, \texttt{graphwalks-parents->128k}, \texttt{gsm-8k-(cot)}, \texttt{gsm8k}, \texttt{gsm8k-chat}, \texttt{hiddenmath}, \texttt{hmmt-2025}, \texttt{hmmt-feb-26}, \texttt{hmmt25}, \texttt{humanity's-last-exam}, \texttt{hypersim}, \texttt{imo-answerbench}, \texttt{intergps}, \texttt{ipho-2025}, \texttt{livebench}, \texttt{livebench-20241125}, \texttt{math}, \texttt{math-(cot)}, \texttt{math-500}, \texttt{matharena-apex}, \texttt{mathverse-mini}, \texttt{mathvision}, \texttt{mathvista}, \texttt{mathvista-mini}, \texttt{mgsm}, \texttt{mmlu}, \texttt{mmlu-(cot)}, \texttt{mmlu-base}, \texttt{mmlu-chat}, \texttt{mmlu-french}, \texttt{mmlu-pro}, \texttt{mmlu-prox}, \texttt{mmlu-redux}, \texttt{mmlu-stem}, \texttt{mmmlu}, \texttt{mmvet}, \texttt{mmvetgpt4turbo}, \texttt{multilingual-mgsm-(cot)}, \texttt{mvbench}, \texttt{nuscene}, \texttt{objectron}, \texttt{olympiadbench}, \texttt{omnimath}, \texttt{openai-mmlu}, \texttt{perceptiontest}, \texttt{phibench}, \texttt{physicsfinals}, \texttt{piqa}, \texttt{pointgrounding}, \texttt{polymath}, \texttt{polymath-en}, \texttt{realworldqa}, \texttt{refcoco-avg}, \texttt{refspatialbench}, \texttt{robospatialhome}, \texttt{sat-math}, \texttt{scienceqa}, \texttt{screenspot}, \texttt{screenspot-pro}, \texttt{stem}, \texttt{sunrgbd}, \texttt{supergpqa}, \texttt{theoremqa}, \texttt{usamo25}, \texttt{we-math}} \\
\midrule
Evaluating Information to Determine Compliance with Standards & 46 & \small{\texttt{agieval}, \texttt{attaq}, \texttt{cybench}, \texttt{cybergym}, \texttt{cybersecurity-ctfs}, \texttt{facts-grounding}, \texttt{factscore}, \texttt{french-mmlu}, \texttt{gdpval-aa}, \texttt{groundui-1k}, \texttt{ifbench}, \texttt{ifeval}, \texttt{longfact-concepts}, \texttt{longfact-objects}, \texttt{lsat}, \texttt{mask}, \texttt{miabench}, \texttt{mmlu}, \texttt{mmlu-(cot)}, \texttt{mmlu-base}, \texttt{mmlu-chat}, \texttt{mmlu-french}, \texttt{mmlu-pro}, \texttt{mmlu-prox}, \texttt{multi-if}, \texttt{openai-mmlu}, \texttt{openbookqa}, \texttt{osworld-g}, \texttt{osworld-screenshot-only}, \texttt{pointgrounding}, \texttt{pope}, \texttt{popqa}, \texttt{refcoco-avg}, \texttt{refspatialbench}, \texttt{screenspot}, \texttt{screenspot-pro}, \texttt{sifo}, \texttt{simpleqa}, \texttt{supergpqa}, \texttt{triviaqa}, \texttt{truthfulqa}, \texttt{uniform-bar-exam}, \texttt{voicebench-avg}, \texttt{wmdp}, \texttt{writingbench}, \texttt{xstest}} \\
\midrule
Getting Information & 81 & \small{\texttt{aa-lcr}, \texttt{bixbench}, \texttt{browsecomp}, \texttt{browsecomp-long-128k}, \texttt{browsecomp-long-256k}, \texttt{browsecomp-vl}, \texttt{browsecomp-zh}, \texttt{complexfuncbench}, \texttt{corpusqa-1m}, \texttt{crag}, \texttt{deepsearchqa}, \texttt{egoschema}, \texttt{facts-grounding}, \texttt{factscore}, \texttt{finsearchcomp-t2-t3}, \texttt{finsearchcomp-t3}, \texttt{flenqa}, \texttt{frames}, \texttt{frontierscience-research}, \texttt{genebench}, \texttt{govreport}, \texttt{graphwalks-bfs->128k}, \texttt{graphwalks-parents->128k}, \texttt{groundui-1k}, \texttt{infinitebench-en.mc}, \texttt{infinitebench-en.qa}, \texttt{longbench-v2}, \texttt{longfact-concepts}, \texttt{longfact-objects}, \texttt{longvideobench}, \texttt{lvbench}, \texttt{mlvu}, \texttt{mm-browsercomp}, \texttt{mmlongbench-doc}, \texttt{mmsearch}, \texttt{mmsearch-plus}, \texttt{mrcr}, \texttt{mrcr-128k-(2-needle)}, \texttt{mrcr-128k-(4-needle)}, \texttt{mrcr-128k-(8-needle)}, \texttt{mrcr-1m}, \texttt{mrcr-1m-(pointwise)}, \texttt{mrcr-64k-(2-needle)}, \texttt{mrcr-64k-(4-needle)}, \texttt{mrcr-64k-(8-needle)}, \texttt{mrcr-v2}, \texttt{mrcr-v2-(8-needle)}, \texttt{natural-questions}, \texttt{nih-multi-needle}, \texttt{nolima}, \texttt{nolima-128k}, \texttt{nolima-32k}, \texttt{nolima-64k}, \texttt{nq}, \texttt{openai-mrcr:-2-needle-128k}, \texttt{openai-mrcr:-2-needle-1m}, \texttt{openai-mrcr:-2-needle-256k}, \texttt{openbookqa}, \texttt{osworld-g}, \texttt{osworld-screenshot-only}, \texttt{pointgrounding}, \texttt{popqa}, \texttt{qasper}, \texttt{qmsum}, \texttt{refcoco-avg}, \texttt{refspatialbench}, \texttt{repoqa}, \texttt{ruler}, \texttt{ruler-1000k}, \texttt{ruler-128k}, \texttt{ruler-2048k}, \texttt{ruler-512k}, \texttt{ruler-64k}, \texttt{screenspot}, \texttt{screenspot-pro}, \texttt{seal-0}, \texttt{simpleqa}, \texttt{squality}, \texttt{summscreenfd}, \texttt{triviaqa}, \texttt{widesearch}} \\
\midrule
Identifying Objects, Actions, and Events & 137 & \small{\texttt{activitynet}, \texttt{ai2d}, \texttt{aitz-em}, \texttt{android-control-high-em}, \texttt{android-control-low-em}, \texttt{androidworld}, \texttt{androidworld-sr}, \texttt{arc-agi}, \texttt{arc-agi-v2}, \texttt{arkitscenes}, \texttt{babyvision}, \texttt{blink}, \texttt{blueprint-bench-2}, \texttt{browsecomp-vl}, \texttt{cc-ocr}, \texttt{charadessta}, \texttt{chartqa}, \texttt{charxiv-d}, \texttt{charxiv-r}, \texttt{countbench}, \texttt{design2code}, \texttt{docvqa}, \texttt{docvqatest}, \texttt{dynamath}, \texttt{egoschema}, \texttt{embspatialbench}, \texttt{erqa}, \texttt{flame-vlm-code}, \texttt{gdpval-mm}, \texttt{graphwalks-bfs-<128k}, \texttt{graphwalks-bfs->128k}, \texttt{graphwalks-parents-<128k}, \texttt{graphwalks-parents->128k}, \texttt{groundui-1k}, \texttt{hallusion-bench}, \texttt{humanity's-last-exam}, \texttt{hypersim}, \texttt{imagemining}, \texttt{infographicsqa}, \texttt{infovqa}, \texttt{infovqatest}, \texttt{intergps}, \texttt{lingoqa}, \texttt{longvideobench}, \texttt{lvbench}, \texttt{mathverse-mini}, \texttt{mathvision}, \texttt{mathvista}, \texttt{mathvista-mini}, \texttt{meld}, \texttt{miabench}, \texttt{mlvu}, \texttt{mlvu-m}, \texttt{mm-browsercomp}, \texttt{mm-if-eval}, \texttt{mm-mind2web}, \texttt{mm-mt-bench}, \texttt{mmau}, \texttt{mmau-music}, \texttt{mmau-sound}, \texttt{mmau-speech}, \texttt{mmbench}, \texttt{mmbench-v1.1}, \texttt{mmbench-video}, \texttt{mme-realworld}, \texttt{mmlongbench-doc}, \texttt{mmmu}, \texttt{mmmu-(val)}, \texttt{mmmu-(validation)}, \texttt{mmmu-pro}, \texttt{mmmuval}, \texttt{mmsearch}, \texttt{mmsearch-plus}, \texttt{mmstar}, \texttt{mmt-bench}, \texttt{mmvet}, \texttt{mmvetgpt4turbo}, \texttt{mmvu}, \texttt{mobileminiwob++-sr}, \texttt{motionbench}, \texttt{mtvqa}, \texttt{muirbench}, \texttt{musiccaps}, \texttt{mvbench}, \texttt{nuscene}, \texttt{objectron}, \texttt{ocrbench}, \texttt{ocrbench-v2}, \texttt{ocrbench-v2-(en)}, \texttt{ocrbench-v2-(zh)}, \texttt{olympiadbench}, \texttt{omnibench}, \texttt{omnibench-music}, \texttt{omnidocbench-1.5}, \texttt{omnigaia}, \texttt{osworld}, \texttt{osworld-extended}, \texttt{osworld-g}, \texttt{osworld-screenshot-only}, \texttt{osworld-verified}, \texttt{perceptiontest}, \texttt{pointgrounding}, \texttt{polymath}, \texttt{pope}, \texttt{qwenwebbench}, \texttt{realworldqa}, \texttt{refcoco-avg}, \texttt{refspatialbench}, \texttt{robospatialhome}, \texttt{scienceqa}, \texttt{scienceqa-visual}, \texttt{screenspot}, \texttt{screenspot-pro}, \texttt{simplevqa}, \texttt{stem}, \texttt{sunrgbd}, \texttt{swe-bench-multimodal}, \texttt{textvqa}, \texttt{tir-bench}, \texttt{v-star}, \texttt{vatex}, \texttt{vcr-en-easy}, \texttt{vibe-eval}, \texttt{video-mme}, \texttt{videommmu}, \texttt{vision2web}, \texttt{visualwebbench}, \texttt{visulogic}, \texttt{vlmsareblind}, \texttt{vqav2}, \texttt{vqav2-(test)}, \texttt{vqav2-(val)}, \texttt{we-math}, \texttt{webvoyager}, \texttt{worldvqa}, \texttt{zerobench}, \texttt{zerobench-sub}} \\
\midrule
Inspecting Equipment, Structures, or Materials & 107 & \small{\texttt{activitynet}, \texttt{ai2d}, \texttt{androidworld}, \texttt{arc-agi}, \texttt{arc-agi-v2}, \texttt{arkitscenes}, \texttt{babyvision}, \texttt{blink}, \texttt{browsecomp-vl}, \texttt{cc-ocr}, \texttt{charadessta}, \texttt{chartqa}, \texttt{charxiv-d}, \texttt{charxiv-r}, \texttt{countbench}, \texttt{design2code}, \texttt{docvqa}, \texttt{docvqatest}, \texttt{dynamath}, \texttt{egoschema}, \texttt{embspatialbench}, \texttt{erqa}, \texttt{flame-vlm-code}, \texttt{groundui-1k}, \texttt{hallusion-bench}, \texttt{humanity's-last-exam}, \texttt{hypersim}, \texttt{imagemining}, \texttt{infographicsqa}, \texttt{infovqa}, \texttt{infovqatest}, \texttt{lingoqa}, \texttt{longvideobench}, \texttt{lvbench}, \texttt{mathverse-mini}, \texttt{mathvision}, \texttt{mathvista}, \texttt{mathvista-mini}, \texttt{miabench}, \texttt{mlvu}, \texttt{mmbench}, \texttt{mmbench-v1.1}, \texttt{mmbench-video}, \texttt{mme-realworld}, \texttt{mmlongbench-doc}, \texttt{mmmu}, \texttt{mmmu-(val)}, \texttt{mmmu-(validation)}, \texttt{mmmu-pro}, \texttt{mmmuval}, \texttt{mmstar}, \texttt{mmt-bench}, \texttt{mmvet}, \texttt{mmvetgpt4turbo}, \texttt{mmvu}, \texttt{motionbench}, \texttt{mtvqa}, \texttt{muirbench}, \texttt{mvbench}, \texttt{nuscene}, \texttt{objectron}, \texttt{ocrbench}, \texttt{ocrbench-v2}, \texttt{ocrbench-v2-(en)}, \texttt{ocrbench-v2-(zh)}, \texttt{olympiadbench}, \texttt{omnibench}, \texttt{omnidocbench-1.5}, \texttt{osworld}, \texttt{osworld-g}, \texttt{osworld-screenshot-only}, \texttt{osworld-verified}, \texttt{perceptiontest}, \texttt{pointgrounding}, \texttt{polymath}, \texttt{pope}, \texttt{realworldqa}, \texttt{refcoco-avg}, \texttt{refspatialbench}, \texttt{robospatialhome}, \texttt{scienceqa}, \texttt{scienceqa-visual}, \texttt{screenspot}, \texttt{screenspot-pro}, \texttt{simplevqa}, \texttt{stem}, \texttt{sunrgbd}, \texttt{swe-bench-multimodal}, \texttt{textvqa}, \texttt{v-star}, \texttt{vatex}, \texttt{vcr-en-easy}, \texttt{vibe-eval}, \texttt{video-mme}, \texttt{videommmu}, \texttt{vision2web}, \texttt{visualwebbench}, \texttt{visulogic}, \texttt{vlmsareblind}, \texttt{vqav2}, \texttt{vqav2-(test)}, \texttt{vqav2-(val)}, \texttt{we-math}, \texttt{webvoyager}, \texttt{worldvqa}, \texttt{zerobench}, \texttt{zerobench-sub}} \\
\midrule
Interpreting the Meaning of Information for Others & 86 & \small{\texttt{alignbench}, \texttt{bbh}, \texttt{big-bench}, \texttt{big-bench-extra-hard}, \texttt{big-bench-hard}, \texttt{boolq}, \texttt{c-eval}, \texttt{charadessta}, \texttt{cluewsc}, \texttt{cmmlu}, \texttt{collie}, \texttt{common-voice-15}, \texttt{commonsenseqa}, \texttt{covost2}, \texttt{covost2-en-zh}, \texttt{csimpleqa}, \texttt{eclektic}, \texttt{fleurs}, \texttt{french-mmlu}, \texttt{giantsteps-tempo}, \texttt{global-mmlu}, \texttt{global-mmlu-lite}, \texttt{govreport}, \texttt{include}, \texttt{lingoqa}, \texttt{maxife}, \texttt{mega-mlqa}, \texttt{mega-tydi-qa}, \texttt{mega-udpos}, \texttt{mega-xcopa}, \texttt{mega-xstorycloze}, \texttt{meld}, \texttt{mm-mt-bench}, \texttt{mmau}, \texttt{mmau-music}, \texttt{mmau-sound}, \texttt{mmau-speech}, \texttt{mmlu}, \texttt{mmlu-(cot)}, \texttt{mmlu-base}, \texttt{mmlu-chat}, \texttt{mmlu-french}, \texttt{mmlu-pro}, \texttt{mmlu-prox}, \texttt{mmlu-redux}, \texttt{mmmlu}, \texttt{mt-bench}, \texttt{multi-if}, \texttt{multichallenge}, \texttt{multilf}, \texttt{multilingual-mmlu}, \texttt{multipl-e}, \texttt{multipl-e-humaneval}, \texttt{musiccaps}, \texttt{nova-63}, \texttt{omnibench-music}, \texttt{open-rewrite}, \texttt{openai-mmlu}, \texttt{qmsum}, \texttt{social-iqa}, \texttt{spider}, \texttt{squality}, \texttt{summscreenfd}, \texttt{superglue}, \texttt{tau-bench-airline}, \texttt{tau-bench-retail}, \texttt{tau2-airline}, \texttt{tau2-retail}, \texttt{tau2-telecom}, \texttt{tau3-telecom}, \texttt{tldr9+-(test)}, \texttt{translation-en→set1-comet22}, \texttt{translation-en→set1-spbleu}, \texttt{translation-set1→en-comet22}, \texttt{translation-set1→en-spbleu}, \texttt{tydiqa}, \texttt{vatex}, \texttt{vocalsound}, \texttt{voicebench-avg}, \texttt{vqav2-(val)}, \texttt{wild-bench}, \texttt{winogrande}, \texttt{wmt23}, \texttt{wmt24++}, \texttt{writingbench}, \texttt{xlsum-english}} \\
\midrule
Judging the Qualities of Objects, Services, or People & 22 & \small{\texttt{acebench}, \texttt{bfcl-v3}, \texttt{crag}, \texttt{finance-agent}, \texttt{finqa}, \texttt{finsearchcomp-t2-t3}, \texttt{finsearchcomp-t3}, \texttt{french-mmlu}, \texttt{gdpval-aa}, \texttt{gdpval-mm}, \texttt{mmlu}, \texttt{mmlu-(cot)}, \texttt{mmlu-base}, \texttt{mmlu-chat}, \texttt{mmlu-french}, \texttt{mmlu-pro}, \texttt{mmlu-prox}, \texttt{openai-mmlu}, \texttt{supergpqa}, \texttt{theoremqa}, \texttt{truthfulqa}, \texttt{writingbench}} \\
\midrule
Making Decisions and Solving Problems & 303 & \small{\texttt{aa-index}, \texttt{aa-lcr}, \texttt{acebench}, \texttt{agieval}, \texttt{ai2-reasoning-challenge-(arc)}, \texttt{ai2d}, \texttt{aider}, \texttt{aime}, \texttt{aime-2024}, \texttt{aime-2025}, \texttt{aime-2026}, \texttt{aitz-em}, \texttt{alignbench}, \texttt{alpacaeval-2.0}, \texttt{amc-2022-23}, \texttt{android-control-high-em}, \texttt{android-control-low-em}, \texttt{androidworld}, \texttt{androidworld-sr}, \texttt{apex-agents}, \texttt{api-bank}, \texttt{arc}, \texttt{arc-agi}, \texttt{arc-agi-v2}, \texttt{arc-c}, \texttt{arc-e}, \texttt{arena-hard}, \texttt{arena-hard-v2}, \texttt{artificial-analysis}, \texttt{autologi}, \texttt{babyvision}, \texttt{bbh}, \texttt{beyond-aime}, \texttt{bfcl}, \texttt{bfcl-v2}, \texttt{bfcl-v3}, \texttt{bfcl-v3-multiturn}, \texttt{bfcl-v4}, \texttt{big-bench}, \texttt{big-bench-extra-hard}, \texttt{big-bench-hard}, \texttt{bixbench}, \texttt{blink}, \texttt{blueprint-bench-2}, \texttt{boolq}, \texttt{browsecomp}, \texttt{browsecomp-long-128k}, \texttt{browsecomp-long-256k}, \texttt{browsecomp-vl}, \texttt{browsecomp-zh}, \texttt{cbnsl}, \texttt{cc-bench-v2-repo}, \texttt{chartqa}, \texttt{charxiv-d}, \texttt{charxiv-r}, \texttt{claw-eval}, \texttt{cluewsc}, \texttt{cmmlu}, \texttt{codeforces}, \texttt{collie}, \texttt{commonsenseqa}, \texttt{complexfuncbench}, \texttt{corpusqa-1m}, \texttt{countbench}, \texttt{crag}, \texttt{cybench}, \texttt{cybergym}, \texttt{deep-planning}, \texttt{deepsearchqa}, \texttt{drop}, \texttt{dynamath}, \texttt{eclektic}, \texttt{egoschema}, \texttt{eq-bench}, \texttt{erqa}, \texttt{evalplus}, \texttt{facts-grounding}, \texttt{finance-agent}, \texttt{finqa}, \texttt{finsearchcomp-t2-t3}, \texttt{finsearchcomp-t3}, \texttt{flenqa}, \texttt{frames}, \texttt{french-mmlu}, \texttt{frontiermath}, \texttt{frontierscience-research}, \texttt{fullstackbench-en}, \texttt{fullstackbench-zh}, \texttt{functionalmath}, \texttt{gdpval-aa}, \texttt{gdpval-mm}, \texttt{genebench}, \texttt{global-mmlu}, \texttt{global-mmlu-lite}, \texttt{global-piqa}, \texttt{gorilla-benchmark-api-bench}, \texttt{gpqa}, \texttt{graphwalks-bfs-<128k}, \texttt{graphwalks-bfs->128k}, \texttt{graphwalks-parents-<128k}, \texttt{graphwalks-parents->128k}, \texttt{gsm-8k-(cot)}, \texttt{gsm8k}, \texttt{gsm8k-chat}, \texttt{hallusion-bench}, \texttt{hellaswag}, \texttt{hiddenmath}, \texttt{hmmt-feb-26}, \texttt{humaneval}, \texttt{humaneval-plus}, \texttt{humanity's-last-exam}, \texttt{imagemining}, \texttt{imo-answerbench}, \texttt{ipho-2025}, \texttt{lingoqa}, \texttt{livebench}, \texttt{livebench-20241125}, \texttt{livecodebench}, \texttt{livecodebench-pro}, \texttt{lmarena-text}, \texttt{longbench-v2}, \texttt{lsat}, \texttt{mask}, \texttt{math}, \texttt{math-(cot)}, \texttt{math-500}, \texttt{matharena-apex}, \texttt{mbpp-plus}, \texttt{mcp-atlas}, \texttt{mcp-mark}, \texttt{mcp-universe}, \texttt{mega-mlqa}, \texttt{mega-tydi-qa}, \texttt{mega-xcopa}, \texttt{mega-xstorycloze}, \texttt{mewc}, \texttt{mgsm}, \texttt{mle-bench-lite}, \texttt{mm-browsercomp}, \texttt{mm-clawbench}, \texttt{mm-if-eval}, \texttt{mm-mind2web}, \texttt{mmau}, \texttt{mmau-music}, \texttt{mmau-sound}, \texttt{mmau-speech}, \texttt{mmbench}, \texttt{mmbench-v1.1}, \texttt{mmbench-video}, \texttt{mmlu}, \texttt{mmlu-(cot)}, \texttt{mmlu-base}, \texttt{mmlu-chat}, \texttt{mmlu-french}, \texttt{mmlu-pro}, \texttt{mmlu-prox}, \texttt{mmlu-redux}, \texttt{mmlu-stem}, \texttt{mmmlu}, \texttt{mmmu}, \texttt{mmmu-(val)}, \texttt{mmmu-(validation)}, \texttt{mmmu-pro}, \texttt{mmmuval}, \texttt{mmsearch}, \texttt{mmsearch-plus}, \texttt{mmstar}, \texttt{mmt-bench}, \texttt{mmvet}, \texttt{mmvetgpt4turbo}, \texttt{mmvu}, \texttt{mobileminiwob++-sr}, \texttt{motionbench}, \texttt{mrcr}, \texttt{mrcr-128k-(2-needle)}, \texttt{mrcr-128k-(4-needle)}, \texttt{mrcr-128k-(8-needle)}, \texttt{mrcr-1m}, \texttt{mrcr-1m-(pointwise)}, \texttt{mrcr-64k-(2-needle)}, \texttt{mrcr-64k-(4-needle)}, \texttt{mrcr-64k-(8-needle)}, \texttt{mrcr-v2}, \texttt{mrcr-v2-(8-needle)}, \texttt{mt-bench}, \texttt{muirbench}, \texttt{multi-if}, \texttt{multi-swe-bench}, \texttt{multichallenge}, \texttt{multilingual-mgsm-(cot)}, \texttt{multilingual-mmlu}, \texttt{musr}, \texttt{mvbench}, \texttt{natural-questions}, \texttt{nexus}, \texttt{nl2repo}, \texttt{nmos}, \texttt{nolima}, \texttt{nolima-128k}, \texttt{nolima-32k}, \texttt{nolima-64k}, \texttt{nq}, \texttt{nuscene}, \texttt{officeqa-pro}, \texttt{ojbench-cpp}, \texttt{olympiadbench}, \texttt{omnibench}, \texttt{omnidocbench-1.5}, \texttt{omnigaia}, \texttt{omnimath}, \texttt{openai-mmlu}, \texttt{openai-mrcr:-2-needle-128k}, \texttt{openai-mrcr:-2-needle-1m}, \texttt{openai-mrcr:-2-needle-256k}, \texttt{openrca}, \texttt{osworld}, \texttt{osworld-extended}, \texttt{osworld-g}, \texttt{osworld-screenshot-only}, \texttt{osworld-verified}, \texttt{paperbench}, \texttt{perceptiontest}, \texttt{phibench}, \texttt{physicsfinals}, \texttt{pinchbench}, \texttt{piqa}, \texttt{polymath}, \texttt{polymath-en}, \texttt{qasper}, \texttt{qwenwebbench}, \texttt{repobench}, \texttt{repoqa}, \texttt{ruler}, \texttt{ruler-1000k}, \texttt{ruler-128k}, \texttt{ruler-2048k}, \texttt{ruler-512k}, \texttt{ruler-64k}, \texttt{sat-math}, \texttt{scienceqa}, \texttt{scienceqa-visual}, \texttt{seal-0}, \texttt{sifo}, \texttt{sifo-multiturn}, \texttt{simpleqa}, \texttt{skillsbench}, \texttt{social-iqa}, \texttt{spider}, \texttt{stem}, \texttt{superglue}, \texttt{supergpqa}, \texttt{swe-bench-multilingual}, \texttt{swe-bench-multimodal}, \texttt{swe-bench-pro}, \texttt{swe-bench-verified}, \texttt{swe-bench-verified-(agentic-coding)}, \texttt{swe-lancer}, \texttt{swe-lancer-(ic-diamond-subset)}, \texttt{t2-bench}, \texttt{tau-bench}, \texttt{tau-bench-airline}, \texttt{tau-bench-retail}, \texttt{tau2-airline}, \texttt{tau2-retail}, \texttt{tau2-telecom}, \texttt{tau3-airline}, \texttt{tau3-banking}, \texttt{tau3-bench}, \texttt{tau3-retail}, \texttt{tau3-telecom}, \texttt{terminal-bench}, \texttt{terminal-bench-2}, \texttt{terminus}, \texttt{theoremqa}, \texttt{tir-bench}, \texttt{toolathlon}, \texttt{truthfulqa}, \texttt{tydiqa}, \texttt{uniform-bar-exam}, \texttt{usamo25}, \texttt{v-star}, \texttt{vcr-en-easy}, \texttt{vending-bench-2}, \texttt{vibe-pro}, \texttt{video-mme}, \texttt{videommmu}, \texttt{visulogic}, \texttt{vita-bench}, \texttt{vlmsareblind}, \texttt{voicebench-avg}, \texttt{vqav2}, \texttt{vqav2-(test)}, \texttt{vqav2-(val)}, \texttt{we-math}, \texttt{webvoyager}, \texttt{widesearch}, \texttt{wild-bench}, \texttt{winogrande}, \texttt{worldvqa}, \texttt{zclawbench}, \texttt{zebralogic}, \texttt{zerobench}, \texttt{zerobench-sub}} \\
\midrule
Monitoring Processes, Materials, or Surroundings & 21 & \small{\texttt{activitynet}, \texttt{attaq}, \texttt{charadessta}, \texttt{cybench}, \texttt{cybergym}, \texttt{cybersecurity-ctfs}, \texttt{embspatialbench}, \texttt{mask}, \texttt{mlvu}, \texttt{mlvu-m}, \texttt{mmbench-video}, \texttt{mmvu}, \texttt{motionbench}, \texttt{mvbench}, \texttt{perceptiontest}, \texttt{pope}, \texttt{robospatialhome}, \texttt{vatex}, \texttt{voicebench-avg}, \texttt{wmdp}, \texttt{xstest}} \\
\midrule
Performing Administrative Activities & 4 & \small{\texttt{finance-agent}, \texttt{gdpval-aa}, \texttt{gdpval-mm}, \texttt{uniform-bar-exam}} \\
\midrule
Performing for or Working Directly with the Public & 3 & \small{\texttt{alignbench}, \texttt{eq-bench}, \texttt{mt-bench}} \\
\midrule
Processing Information & 328 & \small{\texttt{aa-lcr}, \texttt{acebench}, \texttt{agieval}, \texttt{ai2-reasoning-challenge-(arc)}, \texttt{ai2d}, \texttt{aider}, \texttt{aime}, \texttt{aime-2024}, \texttt{aime-2025}, \texttt{aime-2026}, \texttt{aitz-em}, \texttt{alignbench}, \texttt{alpacaeval-2.0}, \texttt{amc-2022-23}, \texttt{android-control-high-em}, \texttt{android-control-low-em}, \texttt{androidworld-sr}, \texttt{apex-agents}, \texttt{api-bank}, \texttt{arc}, \texttt{arc-agi}, \texttt{arc-agi-v2}, \texttt{arc-c}, \texttt{arc-e}, \texttt{arena-hard}, \texttt{arena-hard-v2}, \texttt{autologi}, \texttt{babyvision}, \texttt{bbh}, \texttt{beyond-aime}, \texttt{bfcl}, \texttt{bfcl-v2}, \texttt{bfcl-v3}, \texttt{bfcl-v3-multiturn}, \texttt{big-bench}, \texttt{big-bench-extra-hard}, \texttt{big-bench-hard}, \texttt{bixbench}, \texttt{blink}, \texttt{blueprint-bench-2}, \texttt{boolq}, \texttt{browsecomp}, \texttt{browsecomp-long-128k}, \texttt{browsecomp-long-256k}, \texttt{browsecomp-vl}, \texttt{browsecomp-zh}, \texttt{cbnsl}, \texttt{cc-ocr}, \texttt{charadessta}, \texttt{chartqa}, \texttt{charxiv-d}, \texttt{charxiv-r}, \texttt{cluewsc}, \texttt{cmmlu}, \texttt{codeforces}, \texttt{collie}, \texttt{commonsenseqa}, \texttt{complexfuncbench}, \texttt{corpusqa-1m}, \texttt{countbench}, \texttt{crag}, \texttt{deep-planning}, \texttt{deepsearchqa}, \texttt{design2code}, \texttt{docvqa}, \texttt{docvqatest}, \texttt{drop}, \texttt{dynamath}, \texttt{eclektic}, \texttt{egoschema}, \texttt{eq-bench}, \texttt{erqa}, \texttt{evalplus}, \texttt{facts-grounding}, \texttt{finance-agent}, \texttt{finqa}, \texttt{finsearchcomp-t2-t3}, \texttt{finsearchcomp-t3}, \texttt{flame-vlm-code}, \texttt{flenqa}, \texttt{frames}, \texttt{french-mmlu}, \texttt{frontiermath}, \texttt{frontierscience-research}, \texttt{fullstackbench-en}, \texttt{fullstackbench-zh}, \texttt{functionalmath}, \texttt{gdpval-aa}, \texttt{gdpval-mm}, \texttt{genebench}, \texttt{global-mmlu}, \texttt{global-mmlu-lite}, \texttt{global-piqa}, \texttt{gorilla-benchmark-api-bench}, \texttt{govreport}, \texttt{gpqa}, \texttt{graphwalks-bfs-<128k}, \texttt{graphwalks-bfs->128k}, \texttt{graphwalks-parents-<128k}, \texttt{graphwalks-parents->128k}, \texttt{groundui-1k}, \texttt{gsm-8k-(cot)}, \texttt{gsm8k}, \texttt{gsm8k-chat}, \texttt{hallusion-bench}, \texttt{hellaswag}, \texttt{hiddenmath}, \texttt{hmmt-feb-26}, \texttt{humaneval}, \texttt{humaneval-plus}, \texttt{humanity's-last-exam}, \texttt{if}, \texttt{ifbench}, \texttt{ifeval}, \texttt{imagemining}, \texttt{imo-answerbench}, \texttt{infinitebench-en.mc}, \texttt{infinitebench-en.qa}, \texttt{infographicsqa}, \texttt{infovqa}, \texttt{infovqatest}, \texttt{internal-api-instruction-following-(hard)}, \texttt{ipho-2025}, \texttt{lingoqa}, \texttt{livebench}, \texttt{livebench-20241125}, \texttt{livecodebench}, \texttt{livecodebench-pro}, \texttt{longbench-v2}, \texttt{longvideobench}, \texttt{lsat}, \texttt{lvbench}, \texttt{mask}, \texttt{math}, \texttt{math-(cot)}, \texttt{math-500}, \texttt{matharena-apex}, \texttt{mathverse-mini}, \texttt{mathvision}, \texttt{mathvista}, \texttt{mathvista-mini}, \texttt{mbpp-plus}, \texttt{mcp-atlas}, \texttt{mega-mlqa}, \texttt{mega-tydi-qa}, \texttt{mega-xcopa}, \texttt{mega-xstorycloze}, \texttt{meld}, \texttt{mewc}, \texttt{mgsm}, \texttt{miabench}, \texttt{mlvu}, \texttt{mm-browsercomp}, \texttt{mm-if-eval}, \texttt{mm-mind2web}, \texttt{mm-mt-bench}, \texttt{mmau}, \texttt{mmau-music}, \texttt{mmau-sound}, \texttt{mmau-speech}, \texttt{mmbench}, \texttt{mmbench-v1.1}, \texttt{mmbench-video}, \texttt{mme-realworld}, \texttt{mmlongbench-doc}, \texttt{mmlu}, \texttt{mmlu-(cot)}, \texttt{mmlu-base}, \texttt{mmlu-chat}, \texttt{mmlu-french}, \texttt{mmlu-pro}, \texttt{mmlu-prox}, \texttt{mmlu-redux}, \texttt{mmlu-stem}, \texttt{mmmlu}, \texttt{mmmu}, \texttt{mmmu-(val)}, \texttt{mmmu-(validation)}, \texttt{mmmu-pro}, \texttt{mmmuval}, \texttt{mmsearch}, \texttt{mmsearch-plus}, \texttt{mmstar}, \texttt{mmt-bench}, \texttt{mmvet}, \texttt{mmvetgpt4turbo}, \texttt{mmvu}, \texttt{mobileminiwob++-sr}, \texttt{motionbench}, \texttt{mrcr}, \texttt{mrcr-128k-(2-needle)}, \texttt{mrcr-128k-(4-needle)}, \texttt{mrcr-128k-(8-needle)}, \texttt{mrcr-1m}, \texttt{mrcr-1m-(pointwise)}, \texttt{mrcr-64k-(2-needle)}, \texttt{mrcr-64k-(4-needle)}, \texttt{mrcr-64k-(8-needle)}, \texttt{mrcr-v2}, \texttt{mrcr-v2-(8-needle)}, \texttt{mt-bench}, \texttt{mtvqa}, \texttt{muirbench}, \texttt{multi-if}, \texttt{multi-swe-bench}, \texttt{multichallenge}, \texttt{multilingual-mgsm-(cot)}, \texttt{multilingual-mmlu}, \texttt{musiccaps}, \texttt{musr}, \texttt{mvbench}, \texttt{natural-questions}, \texttt{nih-multi-needle}, \texttt{nolima}, \texttt{nolima-128k}, \texttt{nolima-32k}, \texttt{nolima-64k}, \texttt{nq}, \texttt{nuscene}, \texttt{officeqa-pro}, \texttt{ojbench-cpp}, \texttt{olympiadbench}, \texttt{omnibench}, \texttt{omnibench-music}, \texttt{omnidocbench-1.5}, \texttt{omnigaia}, \texttt{omnimath}, \texttt{openai-mmlu}, \texttt{openai-mrcr:-2-needle-128k}, \texttt{openai-mrcr:-2-needle-1m}, \texttt{openai-mrcr:-2-needle-256k}, \texttt{openrca}, \texttt{osworld}, \texttt{osworld-extended}, \texttt{osworld-g}, \texttt{osworld-screenshot-only}, \texttt{osworld-verified}, \texttt{paperbench}, \texttt{perceptiontest}, \texttt{phibench}, \texttt{physicsfinals}, \texttt{piqa}, \texttt{pointgrounding}, \texttt{polymath}, \texttt{polymath-en}, \texttt{pope}, \texttt{qasper}, \texttt{qmsum}, \texttt{qwenwebbench}, \texttt{repobench}, \texttt{repoqa}, \texttt{ruler}, \texttt{ruler-1000k}, \texttt{ruler-128k}, \texttt{ruler-2048k}, \texttt{ruler-512k}, \texttt{ruler-64k}, \texttt{sat-math}, \texttt{scienceqa}, \texttt{scienceqa-visual}, \texttt{screenspot}, \texttt{screenspot-pro}, \texttt{seal-0}, \texttt{sifo}, \texttt{sifo-multiturn}, \texttt{simpleqa}, \texttt{simplevqa}, \texttt{social-iqa}, \texttt{spider}, \texttt{squality}, \texttt{stem}, \texttt{summscreenfd}, \texttt{superglue}, \texttt{supergpqa}, \texttt{swe-bench-multilingual}, \texttt{swe-bench-multimodal}, \texttt{swe-bench-pro}, \texttt{swe-bench-verified}, \texttt{swe-bench-verified-(agentic-coding)}, \texttt{swe-lancer}, \texttt{swe-lancer-(ic-diamond-subset)}, \texttt{t2-bench}, \texttt{tau-bench}, \texttt{tau-bench-airline}, \texttt{tau-bench-retail}, \texttt{tau2-airline}, \texttt{tau2-retail}, \texttt{tau2-telecom}, \texttt{tau3-airline}, \texttt{tau3-banking}, \texttt{tau3-bench}, \texttt{tau3-retail}, \texttt{tau3-telecom}, \texttt{terminal-bench}, \texttt{terminal-bench-2}, \texttt{terminus}, \texttt{textvqa}, \texttt{theoremqa}, \texttt{tir-bench}, \texttt{toolathlon}, \texttt{truthfulqa}, \texttt{tydiqa}, \texttt{uniform-bar-exam}, \texttt{usamo25}, \texttt{v-star}, \texttt{vatex}, \texttt{vcr-en-easy}, \texttt{vending-bench-2}, \texttt{vibe-eval}, \texttt{video-mme}, \texttt{videommmu}, \texttt{vision2web}, \texttt{visualwebbench}, \texttt{visulogic}, \texttt{vita-bench}, \texttt{vlmsareblind}, \texttt{voicebench-avg}, \texttt{vqav2}, \texttt{vqav2-(test)}, \texttt{vqav2-(val)}, \texttt{we-math}, \texttt{widesearch}, \texttt{wild-bench}, \texttt{winogrande}, \texttt{worldvqa}, \texttt{zebralogic}, \texttt{zerobench}, \texttt{zerobench-sub}} \\
\midrule
Providing Consultation and Advice to Others & 7 & \small{\texttt{finance-agent}, \texttt{finqa}, \texttt{finsearchcomp-t2-t3}, \texttt{finsearchcomp-t3}, \texttt{gdpval-aa}, \texttt{gdpval-mm}, \texttt{uniform-bar-exam}} \\
\midrule
Thinking Creatively & 14 & \small{\texttt{alignbench}, \texttt{alpacaeval-2.0}, \texttt{arena-hard}, \texttt{arena-hard-v2}, \texttt{cc-ocr}, \texttt{collie}, \texttt{creative-writing-v3}, \texttt{eq-bench}, \texttt{meld}, \texttt{mt-bench}, \texttt{mtvqa}, \texttt{open-rewrite}, \texttt{social-iqa}, \texttt{writingbench}} \\
\midrule
Updating and Using Relevant Knowledge & 28 & \small{\texttt{acebench}, \texttt{agieval}, \texttt{french-mmlu}, \texttt{gdpval-aa}, \texttt{gpqa}, \texttt{healthbench}, \texttt{healthbench-hard}, \texttt{lsat}, \texttt{mmlu}, \texttt{mmlu-(cot)}, \texttt{mmlu-base}, \texttt{mmlu-chat}, \texttt{mmlu-french}, \texttt{mmlu-pro}, \texttt{mmlu-prox}, \texttt{mmlu-stem}, \texttt{mmmu}, \texttt{mmmu-(val)}, \texttt{mmmu-(validation)}, \texttt{mmmuval}, \texttt{openai-mmlu}, \texttt{supergpqa}, \texttt{truthfulqa}, \texttt{uniform-bar-exam}, \texttt{videommmu}, \texttt{wmdp}, \texttt{wmt23}, \texttt{writingbench}} \\
\midrule
Working with Computers & 147 & \small{\texttt{acebench}, \texttt{aider}, \texttt{aider-polyglot}, \texttt{aider-polyglot-edit}, \texttt{aitz-em}, \texttt{androidworld}, \texttt{androidworld-sr}, \texttt{apex-agents}, \texttt{api-bank}, \texttt{bfcl}, \texttt{bfcl-v2}, \texttt{bfcl-v3}, \texttt{bfcl-v3-multiturn}, \texttt{bfcl-v4}, \texttt{bigcodebench}, \texttt{bigcodebench-full}, \texttt{bigcodebench-hard}, \texttt{bird-sql-(dev)}, \texttt{bixbench}, \texttt{blueprint-bench-2}, \texttt{browsecomp}, \texttt{browsecomp-vl}, \texttt{cc-bench-v2-backend}, \texttt{cc-bench-v2-frontend}, \texttt{cc-bench-v2-repo}, \texttt{cfeval}, \texttt{claw-eval}, \texttt{codegolf-v2.2}, \texttt{complexfuncbench}, \texttt{crux-o}, \texttt{cruxeval-input-cot}, \texttt{cruxeval-o}, \texttt{cruxeval-output-cot}, \texttt{cybench}, \texttt{cybergym}, \texttt{deep-planning}, \texttt{deepsearchqa}, \texttt{design2code}, \texttt{evalplus}, \texttt{finance-agent}, \texttt{flame-vlm-code}, \texttt{fullstackbench-en}, \texttt{fullstackbench-zh}, \texttt{gdpval-aa}, \texttt{genebench}, \texttt{gorilla-benchmark-api-bench}, \texttt{humaneval}, \texttt{humaneval+}, \texttt{humaneval-average}, \texttt{humaneval-er}, \texttt{humaneval-mul}, \texttt{humaneval-plus}, \texttt{humanevalfim-average}, \texttt{imagemining}, \texttt{instruct-humaneval}, \texttt{lbpp-(v2)}, \texttt{livecodebench}, \texttt{livecodebench-pro}, \texttt{livecodebench-v5}, \texttt{livecodebench-v5-24.12-25.2}, \texttt{livecodebench-v6}, \texttt{mbpp}, \texttt{mbpp+}, \texttt{mbpp-++-base-version}, \texttt{mbpp-evalplus}, \texttt{mbpp-evalplus-(base)}, \texttt{mbpp-pass@1}, \texttt{mbpp-plus}, \texttt{mcp-atlas}, \texttt{mcp-mark}, \texttt{mcp-universe}, \texttt{mewc}, \texttt{mle-bench-lite}, \texttt{mm-browsercomp}, \texttt{mm-clawbench}, \texttt{mm-mind2web}, \texttt{mmsearch}, \texttt{mmsearch-plus}, \texttt{mobileminiwob++-sr}, \texttt{multi-swe-bench}, \texttt{multipl-e-mbpp}, \texttt{natural2code}, \texttt{nexus}, \texttt{nl2repo}, \texttt{octocodingbench}, \texttt{officeqa-pro}, \texttt{ojbench}, \texttt{ojbench-cpp}, \texttt{omnigaia}, \texttt{openrca}, \texttt{osworld}, \texttt{osworld-extended}, \texttt{osworld-g}, \texttt{osworld-screenshot-only}, \texttt{osworld-verified}, \texttt{paperbench}, \texttt{pinchbench}, \texttt{qwenwebbench}, \texttt{repobench}, \texttt{repoqa}, \texttt{seccodebench}, \texttt{sifo}, \texttt{sifo-multiturn}, \texttt{skillsbench}, \texttt{swe-bench-multilingual}, \texttt{swe-bench-multimodal}, \texttt{swe-bench-pro}, \texttt{swe-bench-verified}, \texttt{swe-bench-verified-(agentic-coding)}, \texttt{swe-bench-verified-(agentless)}, \texttt{swe-bench-verified-(multiple-attempts)}, \texttt{swe-lancer}, \texttt{swe-lancer-(ic-diamond-subset)}, \texttt{swe-perf}, \texttt{swe-review}, \texttt{swt-bench}, \texttt{t2-bench}, \texttt{tau-bench}, \texttt{tau-bench-airline}, \texttt{tau-bench-retail}, \texttt{tau2-airline}, \texttt{tau2-retail}, \texttt{tau2-telecom}, \texttt{tau3-airline}, \texttt{tau3-banking}, \texttt{tau3-bench}, \texttt{tau3-retail}, \texttt{tau3-telecom}, \texttt{terminal-bench}, \texttt{terminal-bench-2}, \texttt{terminus}, \texttt{tir-bench}, \texttt{toolathlon}, \texttt{vending-bench-2}, \texttt{vibe}, \texttt{vibe-android}, \texttt{vibe-backend}, \texttt{vibe-ios}, \texttt{vibe-pro}, \texttt{vibe-simulation}, \texttt{vibe-web}, \texttt{vision2web}, \texttt{visualwebbench}, \texttt{vita-bench}, \texttt{webvoyager}, \texttt{widesearch}, \texttt{zclawbench}} \\
\end{longtable}

\phantomsection\label{refs}


\begin{CSLReferences}{1}{0}
\bibitem[\citeproctext]{ref-apra2026ai}
Australian Prudential Regulation Authority. 2026. {``{APRA} Letter to
Industry on Artificial Intelligence ({AI}).''} APRA.
\url{https://www.apra.gov.au/apra-letter-to-industry-on-artificial-intelligence-ai}.

\bibitem[\citeproctext]{ref-asic2024rep798}
Australian Securities and Investments Commission. 2024. {``{REP 798}
Beware the Gap: Governance Arrangements in the Face of {AI}
Innovation.''} ASIC.
\url{https://asic.gov.au/regulatory-resources/find-a-document/reports/rep-798-beware-the-gap-governance-arrangements-in-the-face-of-ai-innovation/}.

\bibitem[\citeproctext]{ref-bis2024ai}
Bank for International Settlements Financial Stability Institute. 2024.
{``Regulating {AI} in the Financial Sector: Recent Developments and Main
Challenges.''} FSI Insights on Policy Implementation 63. Bank for
International Settlements.
\url{https://www.bis.org/fsi/publ/insights63.htm}.

\bibitem[\citeproctext]{ref-bian}
Banking Industry Architecture Network. 2024. {``{BIAN} Service Landscape
14.0.0.''} \url{https://bian.org/servicelandscape-14-0-0/}.

\bibitem[\citeproctext]{ref-yao2025contamination}
Chen, Simin, Yiming Chen, Zexin Li, Yifan Jiang, Zhongwei Wan, Yixin He,
Dezhi Ran, et al. 2025. {``Recent Advances in Large Language Model
Benchmarks Against Data Contamination: From Static to Dynamic
Evaluation.''} \emph{arXiv Preprint arXiv:2502.17521}.
\url{https://arxiv.org/abs/2502.17521}.

\bibitem[\citeproctext]{ref-chiang2024chatbot}
Chiang, Wei-Lin, Lianmin Zheng, Ying Sheng, Anastasios Nikolas
Angelopoulos, Tianle Li, Dacheng Li, Hao Zhang, et al. 2024. {``Chatbot
Arena: An Open Platform for Evaluating {LLMs} by Human Preference.''} In
\emph{Proceedings of the 41st International Conference on Machine
Learning}. \url{https://arxiv.org/abs/2403.04132}.

\bibitem[\citeproctext]{ref-open-llm-leaderboard-v2-2024}
Fourrier, Clémentine, Nathan Habib, Alina Lozada, Kuba Szafer, Thomas
Wolf, Julien Launay, and Edward Beeching. 2024. {``Open {LLM}
Leaderboard V2.''}
\url{https://huggingface.co/spaces/open-llm-leaderboard/open_llm_leaderboard}.

\bibitem[\citeproctext]{ref-eval-harness-2024}
Gao, Leo, Jonathan Tow, Baber Abbasi, Stella Biderman, Sid Black,
Anthony DiPofi, Charles Foster, et al. 2024. {``A Framework for Few-Shot
Language Model Evaluation.''}
\url{https://github.com/EleutherAI/lm-evaluation-harness}.

\bibitem[\citeproctext]{ref-complai2024}
Guldimann, Philipp, Alexander Spiridonov, Robin Staab, Nikola Jovanović,
Mark Vero, Velko Vechev, Anna-Maria Gueorguieva, et al. 2024.
{``{COMPL-AI} Framework: A Technical Interpretation and {LLM}
Benchmarking Suite for the {EU} Artificial Intelligence Act.''}
\emph{arXiv Preprint arXiv:2410.07959}.
\url{https://arxiv.org/abs/2410.07959}.

\bibitem[\citeproctext]{ref-hendrycks2020measuring}
Hendrycks, Dan, Collin Burns, Steven Basart, Andy Zou, Mantas Mazeika,
Dawn Song, and Jacob Steinhardt. 2021. {``Measuring Massive Multitask
Language Understanding.''} \url{https://arxiv.org/abs/2009.03300}.

\bibitem[\citeproctext]{ref-islam2023financebench}
Islam, Pranab, Anand Kannappan, Douwe Kiela, Rebecca Qian, Nino
Scherrer, and Bertie Vidgen. 2023. {``{FinanceBench}: A New Benchmark
for Financial Question Answering.''} \emph{arXiv Preprint
arXiv:2311.11944}. \url{https://arxiv.org/abs/2311.11944}.

\bibitem[\citeproctext]{ref-kiela2021dynabench}
Kiela, Douwe, Max Bartolo, Yixin Nie, Divyansh Kaushik, Atticus Geiger,
Zhengxuan Wu, Bertie Vidgen, et al. 2021. {``Dynabench: Rethinking
Benchmarking in {NLP},''} 4110--24.
\url{https://arxiv.org/abs/2104.14337}.

\bibitem[\citeproctext]{ref-liang2022holistic}
Liang, Percy, Rishi Bommasani, Tony Lee, Dimitris Tsipras, Dilara Soylu,
Michihiro Yasunaga, Yian Zhang, et al. 2023. {``Holistic Evaluation of
Language Models.''} \emph{Transactions on Machine Learning Research}.
\url{https://arxiv.org/abs/2211.09110}.

\bibitem[\citeproctext]{ref-llm-stats-2024}
{``{LLM Stats}: {A}ggregated {LLM} Benchmark Results.''} 2024.
\url{https://llm-stats.com}.

\bibitem[\citeproctext]{ref-onet}
National Center for O*NET Development. 2024. {``{O*NET} Database:
Generalized Work Activities.''} U.S. Department of Labor, Employment and
Training Administration. \url{https://www.onetcenter.org/database.html}.

\bibitem[\citeproctext]{ref-nist2024genai}
National Institute of Standards and Technology. 2024. {``Artificial
Intelligence Risk Management Framework: Generative Artificial
Intelligence Profile ({NIST AI 600-1}).''} NIST.
\url{https://www.nist.gov/publications/artificial-intelligence-risk-management-framework-generative-artificial-intelligence}.

\bibitem[\citeproctext]{ref-yang2024swebenchverified}
OpenAI. 2024. {``Introducing {SWE}-Bench Verified.''}
\url{https://openai.com/index/introducing-swe-bench-verified/}.

\bibitem[\citeproctext]{ref-patil2023gorilla}
Patil, Shishir G, Tianjun Zhang, Xingyao Wang, and Joseph E Gonzalez.
2023. {``Berkeley Function Calling Leaderboard ({BFCL}).''}
\url{https://gorilla.cs.berkeley.edu/blogs/8_berkeley_function_calling_leaderboard.html}.

\bibitem[\citeproctext]{ref-phan2025humanitys}
Phan, Long, Alice Gatti, Ziwen Han, Fan Li, Tianyu Hu, Jeffrey Zhang,
Aliaksei Doroshenko, et al. 2025. {``Humanity's Last Exam.''}
\emph{arXiv Preprint arXiv:2501.14249}.
\url{https://arxiv.org/abs/2501.14249}.

\bibitem[\citeproctext]{ref-rein2024gpqa}
Rein, David, Betty Li Hou, Asa Cooper Stickland, Jackson Petty, Richard
Yuanzhe Pang, Julien Dirani, Julian Michael, and Samuel R Bowman. 2024.
{``{GPQA}: A Graduate-Level Google-Proof q\&a Benchmark.''}
\url{https://arxiv.org/abs/2311.12022}.

\bibitem[\citeproctext]{ref-srivastava2022beyond}
Srivastava, Aarohi, Abhinav Rastogi, Abhishek Rao, Abu Awal Md Shoeb,
Abubakar Abid, Adam Fisch, Adam R Brown, et al. 2023. {``Beyond the
Imitation Game: Quantifying and Extrapolating the Capabilities of
Language Models.''} \emph{Transactions on Machine Learning Research}.
\url{https://arxiv.org/abs/2206.04615}.

\bibitem[\citeproctext]{ref-helm-finance-2024}
Stanford CRFM. 2024. {``{HELM} Finance: Holistic Evaluation of Language
Models on Financial Tasks.''}
\url{https://crfm.stanford.edu/helm/finance/latest/}.

\bibitem[\citeproctext]{ref-suzgun2023challenging}
Suzgun, Mirac, Nathan Scales, Nathanael Schärli, Sebastian Gehrmann, Yi
Tay, Hyung Won Chung, Aakanksha Chowdhery, et al. 2023. {``Challenging
{BIG}-Bench Tasks and Whether Chain-of-Thought Can Solve Them.''}
\url{https://arxiv.org/abs/2210.09261}.

\bibitem[\citeproctext]{ref-wang2019superglue}
Wang, Alex, Yada Pruksachatkun, Nikita Nangia, Amanpreet Singh, Julian
Michael, Felix Hill, Omer Levy, and Samuel R Bowman. 2019.
{``{SuperGLUE}: A Stickier Benchmark for General-Purpose Language
Understanding Systems''} 32. \url{https://arxiv.org/abs/1905.00537}.

\bibitem[\citeproctext]{ref-wang2018glue}
Wang, Alex, Amanpreet Singh, Julian Michael, Felix Hill, Omer Levy, and
Samuel R Bowman. 2019. {``{GLUE}: A Multi-Task Benchmark and Analysis
Platform for Natural Language Understanding.''}
\url{https://arxiv.org/abs/1804.07461}.

\bibitem[\citeproctext]{ref-wang2024mmlupro}
Wang, Yubo, Xueguang Ma, Ge Zhang, Yuansheng Ni, Abhranil Chandra,
Shiguang Guo, Weiming Ren, et al. 2024. {``{MMLU-Pro}: A More Robust and
Challenging Multi-Task Language Understanding Benchmark.''} In
\emph{Advances in Neural Information Processing Systems}. Vol. 37.
\url{https://arxiv.org/abs/2406.01574}.

\bibitem[\citeproctext]{ref-white2025livebench}
White, Colin, Samuel Dooley, Manley Roberts, Arka Pal, Ben Feuer,
Siddhartha Jain, Ravid Shwartz-Ziv, et al. 2025. {``{LiveBench}: A
Challenging, Contamination-Limited {LLM} Benchmark.''} In
\emph{Proceedings of the Thirteenth International Conference on Learning
Representations}. \url{https://arxiv.org/abs/2406.19314}.

\bibitem[\citeproctext]{ref-wu2023bloomberggpt}
Wu, Shijie, Ozan Irsoy, Steven Lu, Vadim Dabravolski, Mark Dredze,
Sebastian Gehrmann, Prabhanjan Kambadur, David Rosenberg, and Gideon
Mann. 2023. {``{BloombergGPT}: A Large Language Model for Finance.''}
\emph{arXiv Preprint arXiv:2303.17564}.
\url{https://arxiv.org/abs/2303.17564}.

\bibitem[\citeproctext]{ref-xie2024finbenneurips}
Xie, Qianqian, Weiguang Han, Zhengyu Chen, Ruoyu Xiang, Xiao Zhang,
Yueru He, Mengxi Xiao, et al. 2024. {``{FinBen}: A Holistic Financial
Benchmark for Large Language Models.''} In \emph{Advances in Neural
Information Processing Systems}. Vol. 37.
\url{https://arxiv.org/abs/2402.12659}.

\bibitem[\citeproctext]{ref-xie2024osworld}
Xie, Tianbao, Danyang Zhang, Jixuan Chen, Xiaochuan Li, Siheng Zhao,
Ruisheng Cao, Toh Jing Hua, Zhoujun Cheng, Dongchan Shi, et al. 2024.
{``{OSWorld}: Benchmarking Multimodal Agents for Open-Ended Tasks in
Real Computer Environments.''} \emph{arXiv Preprint arXiv:2404.07972}.
\url{https://arxiv.org/abs/2404.07972}.

\bibitem[\citeproctext]{ref-xu2024benchmarks}
Xu, Ruijie, Zengzhi Wang, Run-Ze Fan, and Pengfei Liu. 2024.
{``Benchmarking Benchmark Leakage in Large Language Models.''}
\emph{arXiv Preprint arXiv:2404.18824}.
\url{https://arxiv.org/abs/2404.18824}.

\bibitem[\citeproctext]{ref-yao2024taubench}
Yao, Shunyu, Noah Shinn, Pedram Razavi, and Karthik Narasimhan. 2025.
{``\(\tau\)-Bench: A Benchmark for Tool--Agent--User Interaction in
Real-World Domains.''} In \emph{Proceedings of the Thirteenth
International Conference on Learning Representations}.
\url{https://arxiv.org/abs/2406.12045}.

\end{CSLReferences}
\end{document}